\documentclass[12pt]{elsarticle}

\usepackage{url}

\usepackage[table,xcdraw]{xcolor}
\usepackage{amsmath} 
\usepackage{amsfonts}
\usepackage[utf8]{inputenc}

\usepackage{amssymb}
\usepackage{amsmath}

\journal{Biomedical Signal Processing and Control} 
\usepackage[a4paper, total={6in, 8in}]{geometry}

\begin{document}

\begin{frontmatter}

\title{Advancing MRI Reconstruction: A Systematic Review of Deep Learning and Compressed Sensing Integration} 

\author{Mojtaba Safari} 
\author{Zach Eidex} 
\author{Chih-Wei Chang} 
\author{Richard L.J. Qiu} 
\author{Xiaofeng Yang\corref{cor1}} 

\cortext[cor1]{Corresponding author}
\ead{xiaofeng.yang@emory.edu}

%
\affiliation{organization={Department of Radiation Oncology and Winship Cancer Institute, Emory University},
            city={Atlanta},
            postcode={GA 30322}, 
            country={United States of America}}

\begin{abstract}
	Magnetic resonance imaging (MRI) is a non-invasive imaging modality and provides comprehensive anatomical and functional insights into the human body. However, its long acquisition times can lead to patient discomfort, motion artifacts, and limiting real-time applications. To address these challenges, strategies such as parallel imaging have been applied, which utilize multiple receiver coils to speed up the data acquisition process. Additionally, compressed sensing (CS) is a method that facilitates image reconstruction from sparse data, significantly reducing image acquisition time by minimizing the amount of data collection needed. Recently, deep learning (DL) has emerged as a powerful tool for improving MRI reconstruction. It has been integrated with parallel imaging and CS principles to achieve faster and more accurate MRI reconstructions. This review comprehensively examines DL-based techniques for MRI reconstruction. We categorize and discuss various DL-based methods, including end-to-end approaches, unrolled optimization, and federated learning, highlighting their potential benefits. Our systematic review highlights significant contributions and underscores the potential of DL in MRI reconstruction. Additionally, we summarize key results and trends in DL-based MRI reconstruction, including quantitative metrics, the dataset, acceleration factors, and the progress of and research interest in DL techniques over time. Finally, we discuss potential future directions and the importance of DL-based MRI reconstruction in advancing medical imaging. To facilitate further research in this area, we provide a GitHub repository that includes up-to-date DL-based MRI reconstruction publications and public datasets-\url{https://github.com/mosaf/Awesome-DL-based-CS-MRI}.
\end{abstract}

%
%


\end{frontmatter}


\paragraph{\textbf{Abbreviations:}}
\begin{description} 
	\item[bSSFP] Balanced steady-state free precession
	\item[CS] compressed sensing
	\item[DC] Data consistency
	\item[DL] Deep learning
	\item[DL-based MRI reconstruction] Deep learning-based magnetic resonance imaging reconstruction
	\item[DTI] Diffusion-tensor imaging
	\item[E2E] End-to-end
	\item[FFT] Fast Fourier transform
	\item[FID] Fr\'{e}chet inception distance
	\item[FL] Federated learning
	\item[FLAIR] Fluid-attenuated inversion recovery
	\item[GRAPPA] Generalized autocalibrating partially parallel acquisitions
	\item[g-factor] Geometry factor
	\item[GRE] Gradient echo
	\item[MSE] Mean absolute error
	\item[MRA] Magnetic resonance angiography
	\item[MRI] Magnetic resonance imaging
	\item[MSE] Mean square error
	\item[NMSE] Normalized mean square error
	\item[ISTA] Iterative shrinkage-thresholding algorithm
	\item[PD] Proton density
	\item[PDFS] Proton density fat-suppression
	\item[RF] Radio-frequency
	\item[PSNR] Peak signal-to-noise ratio
	\item[R] Acceleration rate
	\item[RMSE] Root mean square error
	\item[RSS] Root sum of squares
	\item[T1c] T1-weighted post-contrast MRI
	\item[T1-MPRAGE] T1-weighted magnetization-prepared rapid gradient-echo
	\item[TV] Total variation
	\item[SENSE] Sensitivity encoding
	\item[SNR] Signal-to-noise ratio
	\item[SSIM] Structural Similarity index
	\item[SWI	] Susceptibility-weighted imaging

\end{description}

\newpage
\section{Introduction}

Magnetic resonance imaging (MRI) is a powerful diagnostic modality widely used for lesion detection and prognosis, radiation treatment planning, and post-treatment follow-up~\cite{https://doi.org/10.1002/jmri.26271, Beigi2018}. It offers high-resolution imaging of soft tissues, making it indispensable for numerous clinical applications. Nonetheless, the Lancet Oncology Commission recently emphasized a critical shortfall of MRI scanners and other imaging technologies in low-income and middle-income countries, contributing to an estimated 2.5 million annual deaths worldwide~\cite{hricak_medical_2021}. This shortfall is partly attributable to the high costs associated with installing, operating, and maintaining MRI systems. As of 2020, there were only seven MRI scanners per million people globally, underscoring the limited availability of this vital technology~\cite{liu_low-cost_2021}.

In addition to financial constraints, lengthy MRI acquisition times hinder the daily throughput of patients and exacerbate wait times~\cite{murali_bringing_2023}. Prolonged scan durations also increase the probability of voluntary and involuntary patient movements, resulting in motion artifacts that degrade image quality~\cite{safari_mri_2023}. Such artifacts not only compromise diagnostic accuracy but also impose a substantial financial burden, with an estimated annual cost of \$364,000 per scanner attributable to motion-related issues~\cite{slipsager_quantifying_2020}.

To address the dual challenges of high cost and extended scan times, advanced MRI reconstruction methods have been developed. In particular, compressed sensing (CS) MRI reconstruction leverages the sparsity of MRI images in certain transform domains to recover high-quality images from undersampled \textit{k}-space data (see Section~\ref{sec:compressed_sensing}). Traditional CS approaches employ randomness in sampling patterns alongside sparsity-enforcing optimization techniques, thereby reducing the number of measurements needed and accelerating the imaging process. By maintaining diagnostic image quality while lowering scan times, CS-based strategies can alleviate financial constraints and increase patient throughput.

In recent years, deep learning (DL) techniques have emerged as a promising complement or alternative to classical CS methods for MRI reconstruction. Unlike CS, which depends on explicit sparsity constraints, DL-based methods learn sophisticated mappings from undersampled to fully sampled images directly from training data (see Section~\ref{sec:deep_learning}). These methods can offer faster reconstruction times and exhibit robust performance in mitigating diverse artifacts and noise. While DL approaches do not explicitly enforce sparsity, their hierarchical structures can implicitly capture sparse representations. Consequently, integrating DL with CS principles can further enhance reconstruction quality and efficiency, thereby bridging the gap between traditional inverse problem formulations and modern data-driven strategies.

Finally, modern MRI scanners largely rely on multi-coil acquisitions, which increase the signal-to-noise ratio (SNR) and enable accelerated imaging through parallel imaging (see Section~\ref{sec:parallel_imaging}) techniques. Although single-coil data commonly serve as a simplified test for newly proposed methods, most studies ultimately validate their techniques on both single-coil and multi-coil datasets to ensure real-world applicability and versatility. This dual-validation approach reflects the practicality and flexibility of current methods in accommodating various acquisition settings.

\subsection{Conventional MRI acceleration techniques}

\subsubsection{Parallel imaging}\label{sec:parallel_imaging}

\begin{figure}
	\centering
	\includegraphics[width=.9\textwidth]{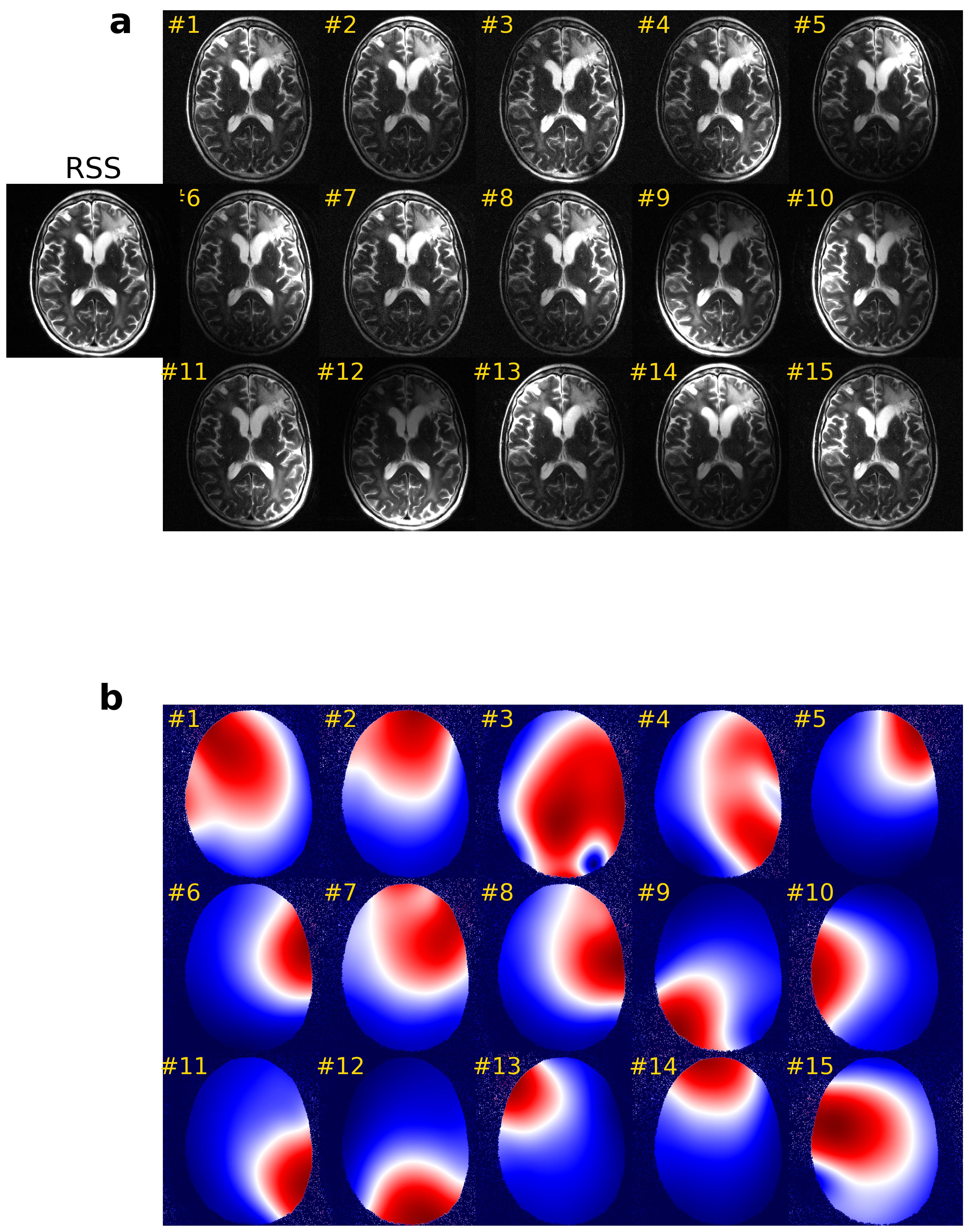}
	\caption{(a) Images and (b) sensitivity maps, estimated by the ESPIRiT approach~\cite{uecker_espiriteigenvalue_2014}, are illustrated for the first 15 receiver coils. Both the images and sensitivity maps are shown as magnitudes. The Root Sum of Squares (RSS) method combines images from multiple receiver coils by computing the square root of the sum of the squares of the individual coil images weighted by the corresponding conjugate sensitivity maps. This combined approach leverages the strengths of both the RSS method and sensitivity weighting to enhance the accuracy of the final reconstructed image.}
	\label{fig:pi_coil_senmap}
\end{figure}

Parallel Imaging (PI) techniques, developed in the late 1990s, have greatly improved MRI by reducing scan times while preserving image quality~\cite{kwiat_decoupled_1991}. PI speeds up image acquisition by using the spatial information from multiple receiver coils. These coils capture the same reduced amount of data, and the acceleration achieved results in undersampled aliased images unless advanced optimization techniques such as generalized autocalibrating partially parallel acquisitions (GRAPPA)~\cite{griswold_generalized_2002} or sensitivity encoding (SENSE)~\cite{pruessmann_sense_1999} are employed. Unlike traditional MRI, which uses a single coil, PI uses an array of coils, each with a unique spatial sensitivity profile (see Figure~\ref{fig:pi_coil_senmap}).

\begin{figure}[tbh!]
	\centering
	\includegraphics[width=\textwidth]{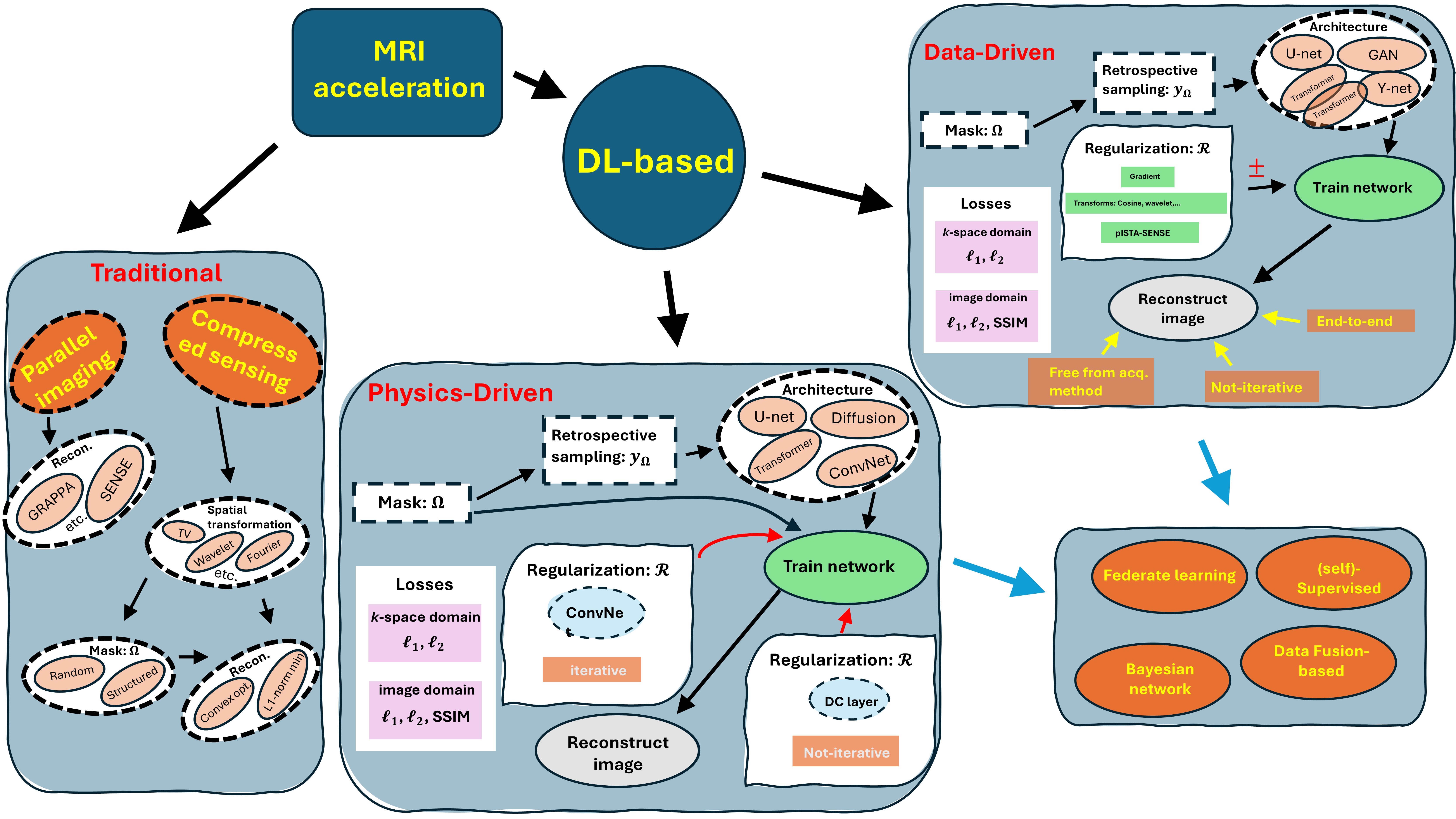}
	\caption{Overview of MRI acceleration techniques: Traditional methods focus on parallel imaging and compressed sensing. Physics-driven approaches incorporate deep learning architectures (e.g., U-net, diffusion models) with regularization and retrospective sampling. Data-driven methods use end-to-end, non-iterative training with architectures like GAN and Y-net. Emerging trends include federated learning, Bayesian networks, and data fusion, advancing MRI reconstruction.}
	\label{fig:overview_all_v2}
\end{figure}

The underlying principle of PI is to use the different spatial sensitivities of multiple coils to reconstruct the missing \textit{k}-space data. This process reduces the number of phase-encoding steps needed, thereby shortening the scan time. The reconstruction algorithms, specific to each PI technique, then use this coil sensitivity information to fill in the gaps in the acquired data~\cite{deshmane_parallel_2012,larkman_parallel_2007}.  There are two fundamental PI techniques as shown in Figure~\ref{fig:overview_all_v2}.

PI is a valuable technique in MRI but has its limitations. One primary limitation is noise amplification, quantified by the geometry factor (g-factor), which increases with higher acceleration factors. The g-factor reflects the noise penalty arising from under-sampling the \textit{k}-space and the subsequent reconstruction process, thereby significantly degrading image quality, particularly in regions with intrinsically low SNR~\cite{pruessmann_sense_1999}. Moreover, PI is susceptible to various artifacts, including aliasing artifacts resulting from under-sampling. Inaccurate estimation of coil sensitivities or limited \textit{k}-space data can produce distorted images, potentially obscuring critical anatomical features and adversely affecting diagnostic confidence~\cite{griswold_generalized_2002}. Consequently, precise calibration of coil sensitivities is critical in accelerated acquisitions, although this process can be labor-intensive. Notably, for fully sampled data, explicit sensitivity maps are not always necessary, as alternative methods such as root-sum-of-squares reconstruction may be used. However, in undersampled scenarios--where the primary goal is to shorten scan times--accurate coil sensitivity information remains crucial to effectively separate aliased signals and maintain robust image fidelity. Changes in the patient’s position or anatomy during the scan can further complicate calibration, emphasizing the need for careful management of coil sensitivities.

While PI can substantially reduce acquisition times, noise amplification, the accuracy of sensitivity maps, and computational constraints collectively limit achievable acceleration factors. As a result, typical clinical implementations utilize acceleration factors (R) of 2 to 4, beyond which the escalation in noise and artifacts can render images diagnostically suboptimal~\cite{lustig_sparse_2007}.

\subsubsection{Compressed sensing}\label{sec:compressed_sensing}

\begin{figure}
	\includegraphics[width=\textwidth]{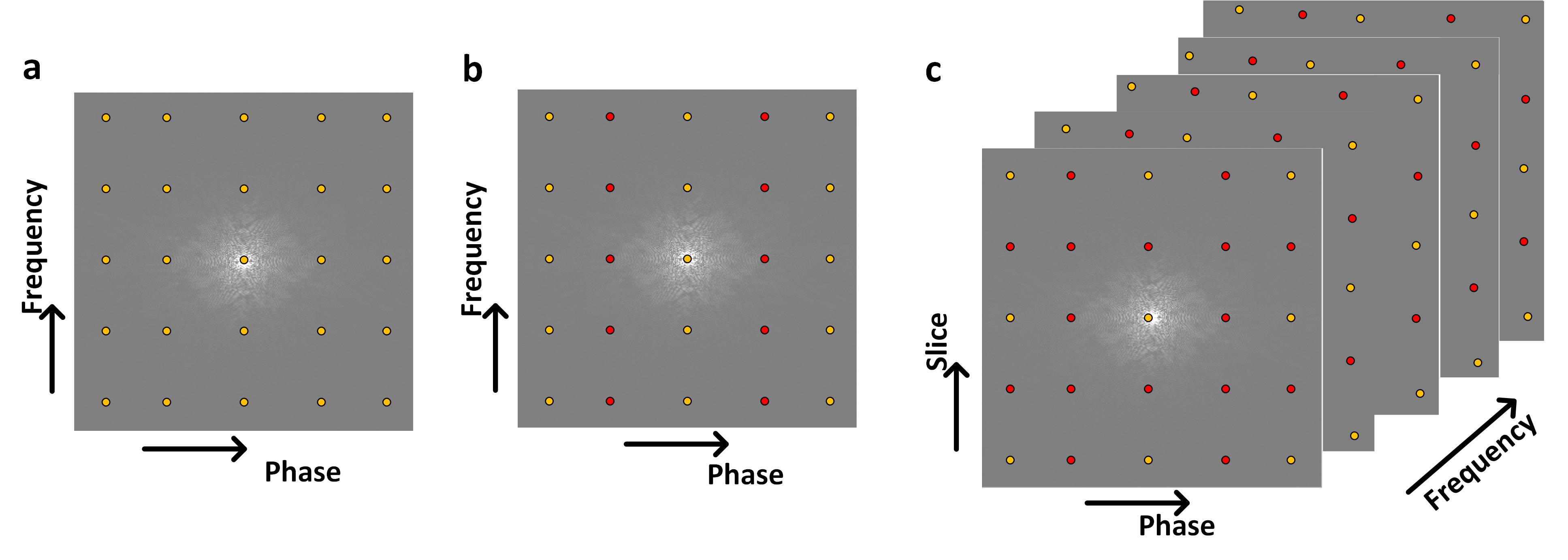}
	\caption{The schematic diagrams of the \textit{k}-space sampling pattern with a Cartesian trajectory are illustrated for (a) fully sampled data and (b) undersampled 2D MRI data in phase encoding direction with acceleration rate (R) $ 2 $ as well as (c) undersampled 3D MRI in both slice and phase encoding direction with $ R=4 $. The yellow and red circles indicate sampled and skipped data during the data acquisition.}
	\label{fig:kspace_sampling}
\end{figure}

MRI requires densely sampled \textit{k}-space to avoid violating the Nyquist criteria~\cite{brown2014chapter12, pipe_2020}, which results in longer acquisition times for high-resolution images. To reduce the imaging time, \textit{k}-space can be undersampled in the phase encoding direction by increasing the spacing between \textit{k}-space lines and, therefore, covering the field of view in a shorter amount of time. Figure~\ref{fig:kspace_sampling} illustrates the \textit{k}-space sampling patterns for fully sampled, 2D undersampled, and 3D undersampled MRI data, demonstrating how CS achieves efficient data acquisition. This is particularly advantageous in MRI, where the acquisition of \textit{k}-space data is inherently time-consuming. CS achieves this by exploiting two main principles: sparsity~\cite{haldar_compressed-sensing_2010, lustig_sparse_2007} and incoherence~\cite{candes_robust_2006}. Figure~\ref{fig:overview_all_v2} illustrates the CS MRI image reconstruction steps.

Reconstructing an image from undersampled data involves solving an optimization problem that enforces both sparsity and data consistency. Let $ y_\Omega \in \mathbb{C}^N$ be the undersampled \textit{k}-space acquired with sampling mask $\Omega$, and let $x \in \mathbb{C}^M$ be the fully sampled data to be recovered, where $N \ll M$ to accelerate MRI acquisition. The corresponding compressed sensing formulation is:

\begin{equation}
	\underset{x}{\arg\min} \parallel y - E_\Omega x \parallel_2 - \beta \parallel \tau (x) \parallel_1
	\label{eq:equation01_cs}
\end{equation}
where $ \tau (\cdot) $ is a sparsity-promoting transform (e.g., wavelet transform), $\beta$ is the regularization parameter. The encoding operator $ E_\Omega = \Omega \mathcal{F} \mathcal{S} $ accounts for the coil sensitivity map $ \mathcal{S} $ and the Fourier transform $ \mathcal{F} $.  This formulation enforces sparsity in the chosen transform domain while ensuring fidelity to the measured multi-coil \textit{k}-space data.

Several iterative algorithms have been developed to solve this optimization problem efficiently. Basis Pursuit solves the $ \ell_1 $-minimization problem directly. The iterative shrinkage-thresholding algorithm (ISTA)~\cite{daubechies_iterative_2004} is an iterative method that alternates between gradient descent and soft-thresholding steps to enforce sparsity. Total variation minimization (TV) enhances edge preservation by minimizing the total variation of the image. However, CS algorithms are unable to completely reconstruct the high-frequency texture content of images~\cite{ravishankar_mr_2010}, limiting them to acceleration factors between 2.5 and 3~\cite{guo_over-and-under_2021}. In addition, these iterative techniques inevitably increase reconstruction time.

\subsubsection{Deep learning-based compressed sensing}\label{sec:deep_learning}

DL has become a transformative tool in the field of MRI, revolutionizing image acquisition, reconstruction, and analysis. By utilizing sophisticated neural network architectures, DL techniques deliver substantial enhancements in image quality, acquisition speed, and diagnostic accuracy, often surpassing conventional methods in various aspects. This section provides an in-depth exploration of the fundamental principles, methodologies, and applications of deep learning-based MRI reconstruction.

For image reconstruction, DL methods have shown the potential to reconstruct high-quality images from undersampled \textit{k}-space data, markedly reducing scan times. Traditional MRI reconstruction methods rely on iterative algorithms that are computationally intensive and time-consuming. In contrast, DL approaches, such as convolutional neural networks (ConvNets), can learn to map reconstructed undersampled \textit{k}-space data (e.g., zero-filled images) to high-quality fully sampled images, effectively suppressing artifacts and restoring missing information. This capability enables rapid and accurate image reconstruction~\cite{zhu_image_2018}. Moreover, DL models have demonstrated superior reconstruction performance, especially at higher acceleration rates, when compared to conventional CS models~\cite{mardani_deep_2018,qin_convolutional_2018}.

Conventional CS relies on explicit sparse representations and iterative optimization to recover images from undersampled data, whereas DL techniques learn relevant features and structures directly from training data. Nevertheless, there is a strong synergy between CS and DL. By combining data fidelity constraints and sampling strategies from CS with the powerful feature extraction and representation capabilities of neural networks, these hybrid methods can achieve superior reconstruction accuracy and speed. 

Many state-of-the-art approaches~\cite{Dratsch2024, 7950457} incorporate elements of both CS and DL through model unrolling or learned regularizers, thereby enforcing sparsity or promoting implicit sparse representations within the network~\cite{hammernik_learning_2018, 10230611, 10.1007/978-3-031-73226-3_10}. In these methods, each iteration of a classical CS optimization algorithm is mapped to one or more layers in a neural network, thereby blending theoretical guarantees from inverse problems with the data-driven adaptability of DL. This integration can help preserve important image details and reduce artifacts while substantially lowering reconstruction times compared to purely iterative CS methods.

Table~\ref{tab:comparative_reviews} provides a concise summary of the most relevant survey articles, revealing technical differences compared to our review. While previous reviews, such as those by \cite{yoon_accelerated_2023} and \cite{chen_ai-based_2022}, have focused on specific aspects of DL-based MRI reconstruction, our review provides a comprehensive analysis that includes detailed explanations of MRI imaging techniques, \textit{k}-space trajectories, sampling patterns, and clinical applications. By categorizing DL methods and providing quantitative metrics and research trends, our review offers a broader and more detailed perspective on the advancements and challenges in this field.

Each study is systematically referenced throughout the manuscript to ensure cohesive integration and enhance the reader’s understanding of how these studies contribute to the broader context of DL-based MRI reconstruction.

\begin{table}[]
	\caption{Related review papers from the DL-based MRI reconstruction.}
	\label{tab:comparative_reviews}
	\resizebox{\textwidth}{!}{%
		\begin{tabular}{p{3cm} p{1cm} p{15cm}}
			\hline
			References                 & Year              &Contributions and technical differences           \\ \hline
			\cite{yoon_accelerated_2023}    & 2023 & This review focuses on accelerated musculoskeletal MRI. It does not discuss the statistics on the quantitative metrics and acceleration rates.\\
			\rowcolor[HTML]{EFEFEF}  \cite{chen_ai-based_2022}    & 2022& This review detailed DL algorithms and provided statistics on quantitative metrics. However, our comprehensive review, in addition to those statistics, provided detailed explanations about MRI imaging, such as \textit{k}-space trajectories, the implications of different sampling patterns, and parallel imaging. Our comprehensive review also discussed the clinical applications of DL-based MRI reconstruction models and provides insights into future directions. \\
			\cite{bustin_compressed-sensing_2020} &  2020          & The primary consideration is DL-based MRI reconstruction models for cardiac imaging. In addition to being wider in scope, our review discusses clinical applications of interest to imaging centers, provides relevant statistics, categorizes the DL method used, and lists related references.  \\
			
			\rowcolor[HTML]{EFEFEF} \cite{xie_review_2022} & 2022 & A review about CS for medical applications. We focus on CS for MRI and categorize it based on the study’s training method. In addition, our comprehensive review provides details about MRI acquisition and acceleration methods.\\
			
			\cite{zeng_review_2021} & 2021 & The main focus of this study is self-supervised DL-based MRI reconstruction algorithms. Our study offers a comprehensive explanation of both traditional methods and deep learning-based compressed sensing MRI algorithms in general. We also have a section dedicated to explaining how these approaches can be trained within the self-supervised framework.\\
			
			\rowcolor[HTML]{EFEFEF} \cite{wang_deep_2021} & 2021 & This review focused on deep learning techniques applied to MR imaging. The review discusses various methods developed from 2016 to June 2020, focusing on accelerating MR imaging through deep learning-based reconstructions from undersampled \textit{k}-space data. However, our study provides more recent training frameworks, such as federated learning, and models, such as conditional diffusion models. However, our study has dedicated sections to discuss traditional methods such as parallel imaging and compressed sensing, \textit{k}-space trajectories, and sampling methods and their clinical applications. In addition, our method provides statistical analyses of the DL-based MRI reconstruction approaches, quantitative metrics, and datasets that were employed. Our study discussed clinical evaluations. \\
			
			\cite{oscanoa_deep_2023} & 2023 & This study focused on DL-based MRI reconstruction models for cardiac imaging, providing relevant statistics and clinical applications. In contrast, our review is broader in scope, covering a wider range of clinical applications beyond cardiac imaging. Our review also categorizes different DL methods, provides a more detailed discussion and statistics of the technical aspects of MRI acquisition and acceleration methods, and includes a broader set of quantitative metrics.\\
			\hline
		\end{tabular}%
		
	}
	
\end{table}

\begin{figure}[tbh!]
	\includegraphics[width=\textwidth]{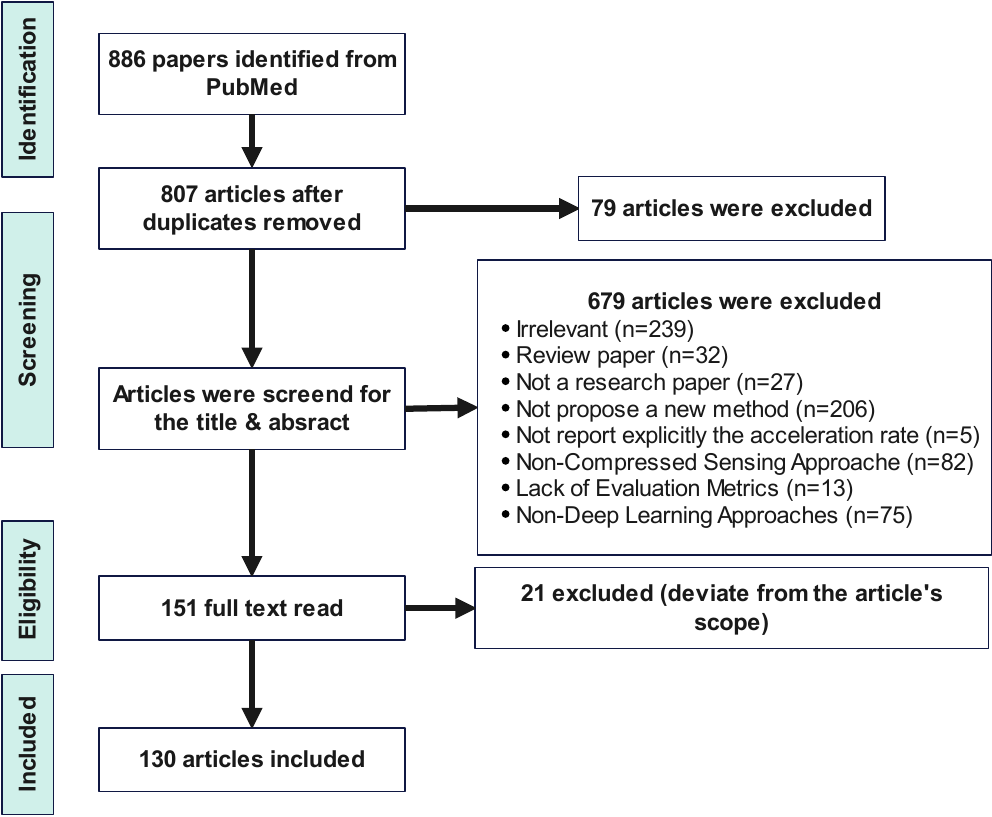}
	\caption{Flowchart of our study selection process.}
	\label{fig:flowchart}
\end{figure}

The PubMed database was meticulously searched on February 1st, 2024 and revised in January 1st 2025, using the terms ``deep learning reconstruction'' OR ``fastMRI'' OR ``unrolled optimization'' OR ``MRI reconstruction'' OR ``MRI acceleration'' for articles published from January 2016 to January 2025. Relevant studies were carefully screened by title and abstract content. Of the 886 publications identified by PubMed, 130 articles were included. Figure~\ref{fig:flowchart} illustrates the entire literature screening and selection process.

\subsection{\textit{k}-space trajectories}
\begin{figure}[tbh!]
	\centering
	\includegraphics[width=\textwidth]{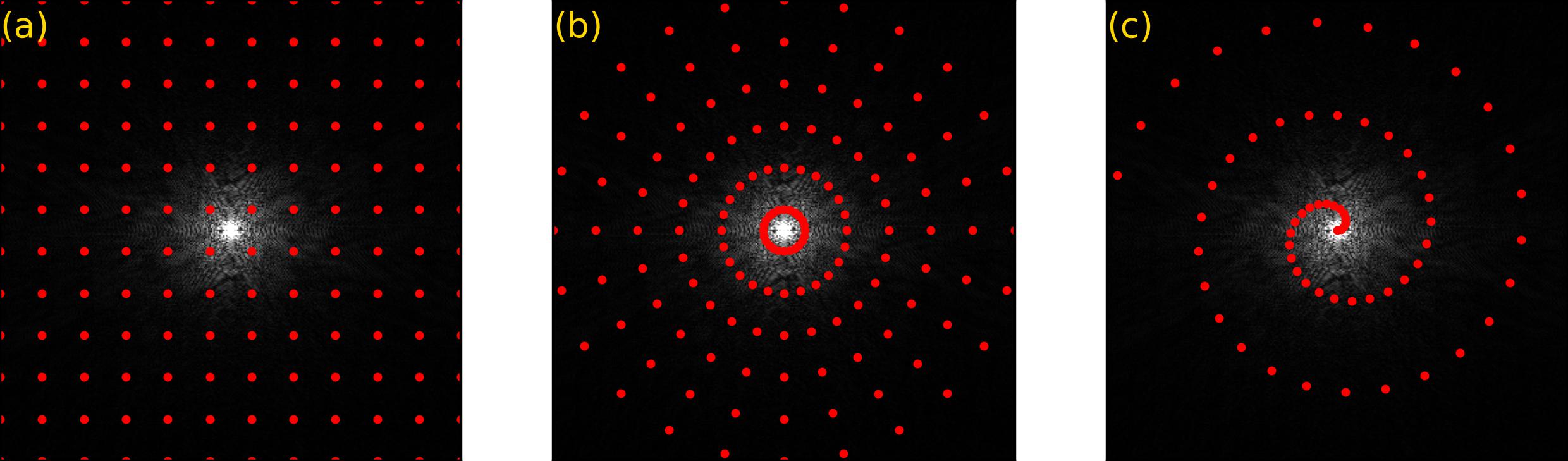}
	\caption{An exemplary \textit{k}-space with overlaid (a) Cartesian, (b) radial, and (c) spiral trajectories illustrated.}
	\label{fig:trajectories}
\end{figure}

Various trajectories have been developed in MRI for traversing \textit{k}-space, including Cartesian, spiral, radial, and random trajectories~\cite{jHennig_1991_re}. The Cartesian trajectory, as depicted in Figure~\ref{fig:trajectories}a, consists of parallel lines equidistant from each other, with each line representing a frequency-encoding readout. The image can be reconstructed using a fast Fourier transform (FFT), but each line requires a separate RF pulse, prolonging imaging time.

The radial trajectory was first used by~\cite{lauterbur_image_1973} and shown in Figure~\ref{fig:trajectories}b, consists of spokes starting from the center, with an oversampling center in \textit{k}-space that makes it robust to motion artifacts~\cite{mcmahan_communication-efficient_2017}. However, under-sampling in the azimuthal direction increases streak artifacts~\cite{xue_automatic_2012}.

The spiral trajectory shown in Figure~\ref{fig:trajectories}c was introduced to decrease the MRI acquisition time. It starts at the center of the \textit{k}-space and spirals outward, similar to radial sampling, and is robust to motion artifacts. However, hardware limitations restrict imaging efficiency and increase image blurring.

\subsubsection{Sampling patterns}

\begin{figure}[tbh!]
	\centering
	\includegraphics[width=\textwidth]{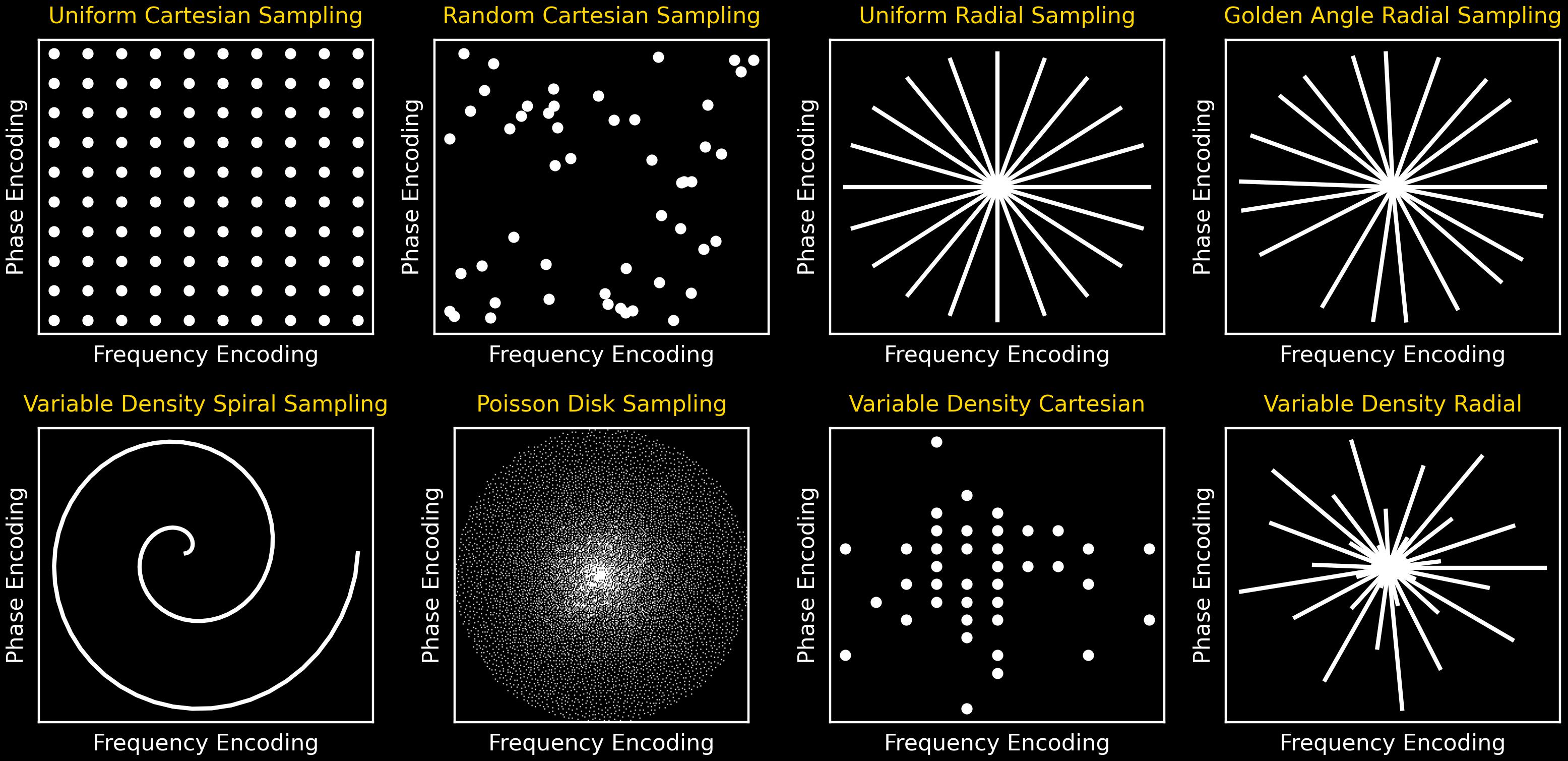}
	\caption{Widely used \textit{k}-space under-sampling patterns are illustrated.}
	\label{fig:sampling_trajectory_v02}
\end{figure}

Many sampling strategies, when integrated with DL, facilitate accelerated MRI without severely compromising image quality~\cite{YIASEMIS202433}. These strategies provide undersampled \textit{k}-space data to DL-based methods, which recover high-fidelity images by exploiting learned representations. Figure~\ref{fig:sampling_trajectory_v02} illustrates some widely used sampling patterns.

A prevalent method for data sampling is uniform Cartesian sampling, where data points are collected following a regularly spaced Cartesian grid. Although straightforward, this technique may cause coherent aliasing artifacts that reduce the efficacy of CS~\cite{safari_mri_2024}. In contrast, random Cartesian sampling randomly places sampled points across the Cartesian grid, resulting in incoherent aliasing, which generally leads to better reconstruction quality~\cite{lustig_sparse_2007}.

While Cartesian sampling is often depicted in a line-based format, it is not exclusively limited to this structure. Variable density Poisson sampling--commonly known as ``Cartesian Poisson'' sampling--randomly selects points from an underlying Cartesian grid according to a specified density profile. This randomness can help distribute aliasing artifacts more uniformly, thereby improving CS reconstructions. Likewise, pseudo-non-Cartesian methods such as CIRCUS~\cite{QIMS3439} may appear to diverge from traditional line-based formats but can still be mapped onto a Cartesian grid. These techniques leverage the straightforwardness of FFT-based reconstruction while utilizing incoherence properties that are crucial to compressed sensing.

Beyond strictly Cartesian formats, radial sampling evenly distributes spokes through \textit{k}-space, allowing for oversampling of the center, which is beneficial for motion robustness~\cite{terpstra_accelerated_2023}. A variant known as golden angle radial sampling positions consecutive radial spokes using the golden angle increment, improving coverage over time for dynamic and real-time imaging~\cite{winkelmann_optimal_2007}. Meanwhile, spiral sampling traverses \textit{k}-space along spiral trajectories, accommodating rapid acquisitions suitable for functional MRI and angiography~\cite{glover_simple_1999}.

Randomization strategies are also significant in Poisson disk sampling, which guarantees that no two sampled points are too closely spaced. This characteristic produces a quasi-uniform yet random arrangement, enhancing incoherence and benefiting various MRI applications under CS~\cite{slavkova_untrained_2023}. Furthermore, k–t sampling, as demonstrated by k–t FOCUSS, combines \textit{k}-space (spatial) and t-space (temporal) sampling to facilitate accelerated dynamic imaging by leveraging spatiotemporal sparsity~\cite{jung_kt_2009}. In both two and three dimensions, non-Cartesian sampling can adopt various forms (e.g., radial, spiral)~\cite{hu_run-up_2021}, with three-dimensional expansions allowing for volumetric imaging and potentially enhancing efficiency.

In recent years, optimized sampling patterns derived from machine learning and advanced optimization algorithms have gained significant importance. By tailoring sampling masks to the unique characteristics of the target images or applications, these methods enhance reconstruction outcomes further~\cite{wang_stochastic_2023}. Ultimately, designing sampling patterns to achieve the necessary incoherence is crucial for CS. When these strategies are integrated with DL, undersampled measurements can often be reconstructed with high fidelity, thereby reducing scan times while maintaining image quality.

\section{Deep learning}
\subsection{Convolutional neural network}

Convolutional neural networks, also known as ConvNets, are a type of deep neural networks that are designed to analyze grid-like data such as images and speech~\cite{lecun_convolutional_1995}. They have gained widespread recognition after the success of AlexNet~\cite{krizhevsky_imagenet_2012} and have since been used to achieve state-of-the-art performance in various medical image processing and analysis tasks. ConvNets typically consist of multiple layers, with each layer including a convolutional operator, batch normalization layers, nonlinear activation functions, and dropout layers. Nonlinear activation functions facilitate the learning of complex functions. Finally, weight regularization and dropout layers mitigate overfitting. During convolution, trainable convolution kernels slide over the images to extract multiple feature maps called channels.

The network’s parameters are computed using a backpropagation algorithm that calculates the gradient of the cost function with respect to the parameters in each layer. (Batch) Normalization layers are crucial in training deep ConvNets to prevent vanishing and exploding gradients. In addition, residual blocks~\cite{he_deep_2016} are a popular choice for building advanced ConvNets due to their ability to prevent gradient vanishing and facilitate smoother error surfaces~\cite{li_visualizing_2018}. By incorporating skip-layer connections between input and output, residual blocks can help reduce the risk of local minima. Furthermore, when combined with (batch) normalization layers, residual blocks can effectively address the problems of vanishing and exploding gradients. 

\subsection{Network architectures}

\subsubsection{U-Net}

Several DL models with different architectures have been proposed to enhance the performance and generalization of ConvNets. Among them, U-Net, with its elegant design that utilizes skip connections between the encoder and decoder, is the best-known architecture in computer vision~\cite{ronneberger_u-net_2015, 10643318}. It has been extensively exploited in different medical applications such as image synthesis~\cite{han_mr-based_2017}, segmentation~\cite{dong_automatic_2019,safari2024information}, and registration~\cite{balakrishnan_voxelmorph_2019}. In recent years, U-net architectures have incorporated residual and attention layers as a backbone to increase the network’s depth and improve performance.

\subsubsection{Transformer architectures}

ConvNets have demonstrated impressive efficacy in various medical imaging tasks. However, they are limited by the local context of convolutional operations limits their ability to capture global contextual information. To overcome this limitation, transformers with long-range dependencies have been developed to capture global context~\cite{dosovitskiy_image_2020}. However, transformer models with more trainable parameters require larger databases for training, which can be a challenge in medical imaging, where data scarcity and computational resources may be constrained.

To address these challenges, various variations have been proposed, such as Swin Transformers~\cite{liu_swin_2021}, Vision CNN-Transformer~\cite{fang_hybrid_2022, eidex2024deep}, and ReconFormer~\cite{guo_reconformer_2023}, which aims to reduce model size while improving or maintaining performance. For instance, Swin Transformers employ hierarchical feature maps and shifted windows to achieve computational efficiency, whereas Vision CNN-Transformer integrates convolutional layers with Transformer blocks to leverage the strengths of both paradigms. ReconFormer is a lightweight recurrent transformer model that iteratively reconstructs high-fidelity MRI images from highly undersampled \textit{k}-space data.

Recent advancements further demonstrate the potential of transformers in MRI reconstruction, exemplified by SLATER, a zero-shot learned adversarial transformer that employs cross-attention blocks to enhance sensitivity to global spatial interactions \cite{korkmaz_slater_2022}. By decoupling the MRI prior from the imaging operator through a combination of generative pretraining and zero-shot optimization, SLATER achieves superior performance across diverse datasets and under-sampling patterns, thereby underscoring the transformative impact of transformer-based models in medical imaging. Ongoing research continues to explore unrolled and hybrid approaches to bolster the adaptability and clinical utility of these architectures~\cite{korkmaz_self_supervised_2023}.

\subsection{Generative learning paradigms}
\subsubsection{Generative adversarial network}

Generative adversarial networks (GANs) are implicit generative models. Unlike explicit generative models, GANs do not attempt to directly define the likelihood function. Instead, they aim to generate samples that are indistinguishable from real data through an adversarial process between a generator and a discriminator. The GAN, initially introduced in 2014, consists of two networks, generative and discriminator~\cite{goodfellow_generative_2020}. The former is trained to generate artificial data samples to approximate the target data distribution, and the latter is simultaneously trained to distinguish the artificial data from real ground truth data. Thus, the discriminators encourage the generator to generate data samples with a distribution similar to the target distribution. Variations of GANs have been developed to perform tasks including image-to-image translation, such as conditional GAN~\cite{mirza_conditional_2014}, StyleGAN~\cite{karras_style-based_2019}, CycleGAN~\cite{zhu_unpaired_2017}, and Pix2Pix~\cite{isola_image--image_2017}. GANs are widely used in medical imaging for tasks such as image registration, image synthesis, MRI image reconstruction, and MRI artifact reduction~\cite{quan_compressed_2018,shaul_subsampled_2020,yang_dagan_2017, safari2024unsupervised}.

\subsubsection{Diffusion model}
\begin{figure}[tbh!]
	\centering
	\includegraphics[width=\textwidth]{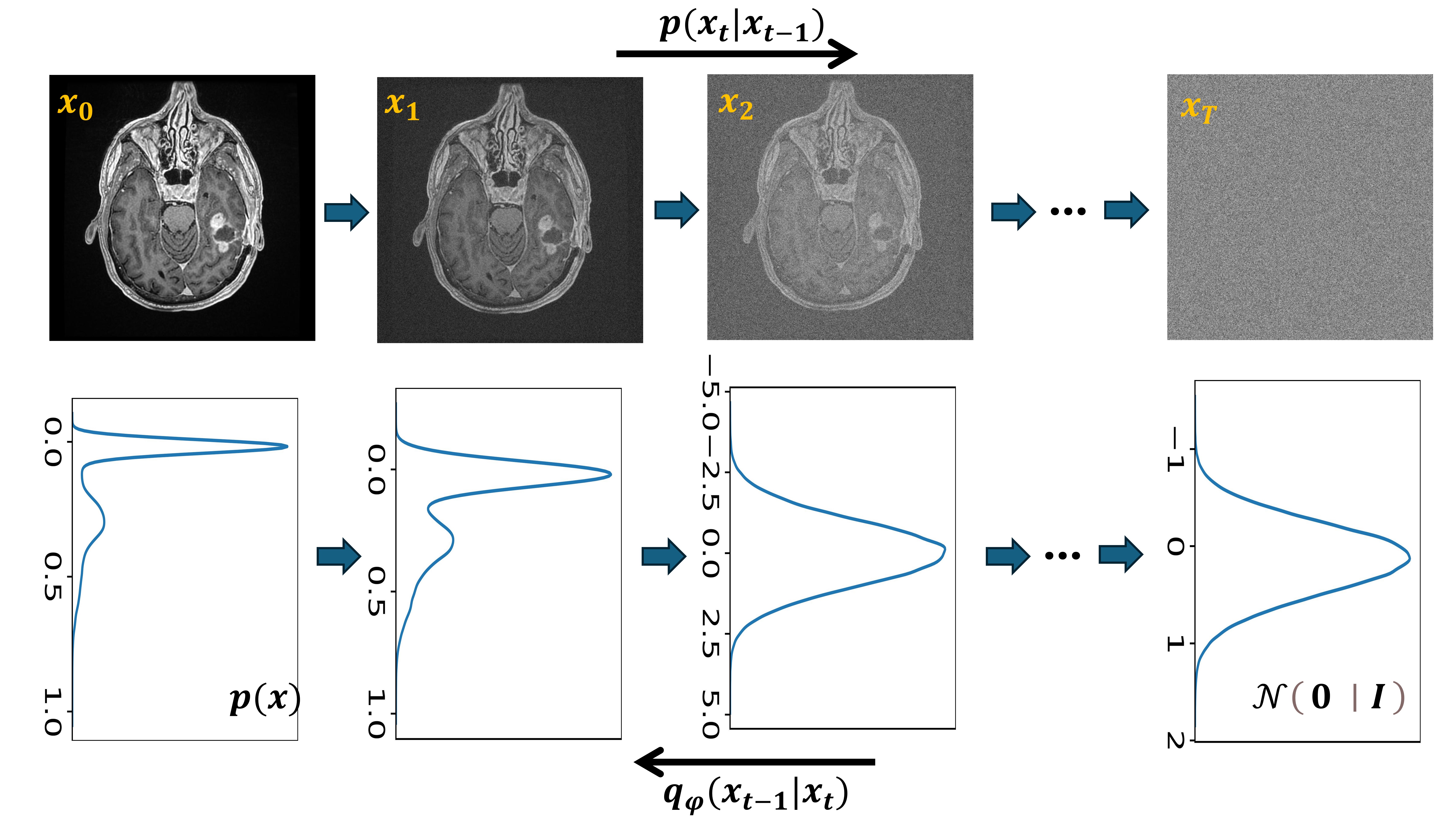}
	\caption{Forward and reverse diffusion processes. The first row indicates data in image space, and the second row indicates the corresponding data distribution. The forward diffusion process modeled by $ p $ adds Gaussian noise in $ T $ steps in a controlled way to produce normal Gaussian noise in step $ T $. The reverse diffusion process modeled by $ q_\varphi $ with learnable parameters $ \varphi $ aim to learn input $ x_0 $ from noise-corrupted image $ x_t $ in a given step t.}
	\label{fig:ddpm_v01}
\end{figure}

The stable diffusion model, inspired by nonequilibrium thermodynamics, aims to simplify complex and difficult-to-calculate distributions using tractable ones like normal Gaussian distributions~\cite{sohl-dickstein_deep_2015}. This model is comprised of two steps - the forward and reverse processes (Figure~\ref{fig:ddpm_v01}). During the forward process, Gaussian noise is added to the initial image $ x_0 $ over $ T $ steps until the data at step $T$ becomes normal Gaussian noise $ x_T\sim \mathcal{N}(\mathbf{0},\mathbf{I}) $. In the reverse process, the model learns to recover the original image $ x_0 $ from its noisy version given at a step $ t\in (0, T] $~\cite{chan_tutorial_2024}.

Diffusion-based reconstruction models represent a significant advancement in MRI reconstruction techniques. This emerging approach shows promise for improving reconstruction accuracy and robustness, particularly in scenarios where traditional methods face limitations. The stable diffusion models have been employed for medical imaging tasks such as denoising~\cite{pan_2d_2023}, synthesis~\cite{pan_synthetic_2023}, MRI distortion reduction~\cite{safari_mri_2023}, and MRI image reconstruction~\cite{chung_score-based_2022, gungor_adaptive_2023, safari2024selfsupervisedadversarialdiffusionmodels}.

While diffusion models provide a robust framework for image reconstruction similar to other generative models like GANs and VAEs, it is important to note that they are not standalone MRI reconstruction methods. Instead, they offer a versatile and powerful approach that can be integrated into the broader MRI reconstruction pipeline, complementing existing techniques and addressing specific challenges.

\section{Deep learning for MRI reconstruction}

The framework for DL-based MRI reconstruction models can be divided into two main approaches: physics-driven and fully data-driven models. While all DL-based methods rely on data for training, physics-driven models incorporate domain-specific knowledge, such as sampling patterns and physical constraints, into the reconstruction process, whereas fully data-driven models leverage the data alone to learn the reconstruction. Within these categories, there are two types of models: end-to-end and unrolled. End-to-end models take in zero-filled \textit{k}-space and output fully sampled \textit{k}-space. They typically utilize a regularization term listed in Table~\ref{tab:table3_reqularizations} to enforce the uniqueness of the reconstructed images. On the other hand, unrolled models are more complex and further classified into two types: unrolled optimization and closed-form models known for the ``data consistency (DC) layer.'' unrolled optimization models iteratively optimize the reconstruction process, while DC layer models use a closed-form equation to ensure data consistency. Table~\ref{tab:table2_comparative} presents a summary comparing DL-based MRI reconstruction methods. These models are utilized in various training scenarios, including federated learning and self-supervised training.

\begin{table}[tbh!]
	\caption{The widely used regularization terms in DL-based MRI reconstruction approaches are summarized.}
	\label{tab:table3_reqularizations}
	\resizebox{\textwidth}{!}{%
		\begin{tabular}{p{4cm} p{4cm} p{4cm}p{4cm}}
			\hline
			Traditional                                                                                                                                                                              & DL-based model                                                                                                                                                                    & Regularizer                                                                                                                                                               & Reference                                                                                                                                                                                             \\\hline
			Dictionary learning                                                                                                                                                                      & ADMS                                                                                                                                                                              &     $   \parallel \alpha \parallel_1    $                & \cite{cao_cs-mri_2020} \\
			Field of expert                                                                                                                                                                          & VN                  & $ \sum_i \langle \Phi_i (K_i x), 1 \rangle  $      & Hammernik \textit{et al.}~\cite{hammernik_learning_2018}  \\
			pISTA-SENSE                                                                                                                                                                              & pISTA-SENSE-ResNet                                                                                                                                                                &   $ \parallel \Psi x \parallel_1 $     & \cite{lu_pfista-sense-resnet_2020} \\
			Total variation                                                                                                                                                                          & -, RELAX                                                                                                                                                                          &   $ \parallel \nabla x \parallel_1 $  & \cite{liu_regularization_2021, sun_compressed_2017} \\
			Sparse and low-rank model                                                                                                                                                                & ODLS                                                                                                                                                                              &     $ \parallel \mathcal{H}x \parallel_* $                    & \cite{wang_one-dimensional_2022} \\ \hline
			\multicolumn{4}{l}{\begin{tabular}[c]{@{}l@{}}
					\textbf{Dictionary learning} learns a latent representation of the input image where the norm enforces\\ it to be sparse.  \\
					
					\textbf{Field of expert} is composed of the convolution kernel and , which they are learned from data. \\ 
					
					\textbf{pISTA-SENSE} is a projected iterative soft-thresholding algorithm that solves (\ref{eq:equation01_cs}) iteratively\\ where the transform enforces the sparsity of the reconstructed image. \\ 
					
					\textbf{Total variation} enforces image smoothness by minimizing the image gradient variations. \\ 
					
					\textbf{Sparse and low-rank models} minimize the nuclear norm $ \parallel \mathcal{H}x \parallel_* $, where $  \mathcal{H} $ is the Hankel\\ matrix.\end{tabular}}
		\end{tabular}
	}
\end{table}

\begin{table}[tbh!]
	\caption{Comparative summary of DL-based MRI reconstruction methods.}
	\label{tab:table2_comparative}
	\resizebox{\textwidth}{!}{%
		\begin{tabular}{p{2cm} p{4cm} p{3cm}p{3cm}p{3cm}p{2cm}}
			\hline
			Method                 & Main idea                                                                                & Improvements                              & Advantages                                       & Disadvantages                                            & Key reference           \\\hline
			End-to-end             & Directly reconstructs fully sampled images from undersampled data using neural networks & Enhances image quality, reduces scan time & Easy to implement, applicable to various domains & Requires large datasets, potential overfitting           & \cite{levac_accelerated_2023} \\
			\rowcolor[HTML]{EFEFEF} unrolled Optimization    & Maps iterative steps of optimization algorithms to neural network layers                 & More efficient reconstructions            & Requires smaller datasets, good generalization   & Increased computation time during training and inference & \cite{aggarwal_modl_2018} \\
			Data Consistency Layer & Ensures fidelity to acquired data using closed-form equations in the network             & Enhances reconstruction accuracy          & Requires smaller datasets, easy to implement     & Requires knowledge of under-sampling pattern             & \cite{qin_convolutional_2018}     \\\hline
		\end{tabular}
	}
\end{table}

The physics-driven approach begins with the application of a mask ($ \Omega $) to the acquired MRI data, representing the sampling pattern used during the MRI acquisition. This mask is essential for creating undersampled \textit{k}-space data, which is referred to as retrospective sampling ($ y_\Omega $). The undersampled data is then fed into various neural network architectures, such as U-net, diffusion models, transformers, and ConvNets. These networks are trained to reconstruct high-quality images from the undersampled data. During the training phase, the network learns regularization ($\mathcal{R} $) using ConvNets and iterative methods. This iterative process ensures that the reconstructed images adhere to expected properties derived from the image acquisition models and improve over successive iterations. The training process is guided by loss functions calculated in both the \textit{k}-space domain ($ \ell_1 $, $ \ell_2 $) and the image domain ($ \ell_1 $, $ \ell_2 $, SSIM).

On the other hand, the data-driven approach also begins with applying a mask ($ \Omega $) to generate undersampled data ($ y_\Omega $). However, this approach relies heavily on advanced architectures like U-net, GANs, Y-net, and transformers for end-to-end image reconstruction. The training process in this approach involves feeding the undersampled data into these networks and employing various regularization techniques. Unlike the physics-driven approach, the data-driven method can utilize non-iterative processes for regularization, such as gradient-based methods, cosine and wavelet transforms, and pISTA-SENSE~\cite{lu_pfista-sense-resnet_2020} algorithms. The loss functions used are similar to those in the physics-driven approach, computed in both \textit{k}-space ($ \ell_1 $ and $ \ell_2 $) and the image domain ($ \ell_1 $, $ \ell_2 $, and SSIM). This approach aims to produce the reconstructed image directly through the network without the need for multiple iterations, making it an end-to-end solution that is free from being tightly coupled with specific data acquisition methods.

Furthermore, Figure~\ref{fig:overview_all_v2} highlights the emerging advanced techniques in DL-based MRI reconstruction models. These include federated learning (FL), which involves training models across multiple decentralized devices holding local data samples without exchanging them (see Section~\ref{sub:federatedLearning}).

\begin{figure}[tbhp!]
	\centering
	\includegraphics[width=.8\textwidth]{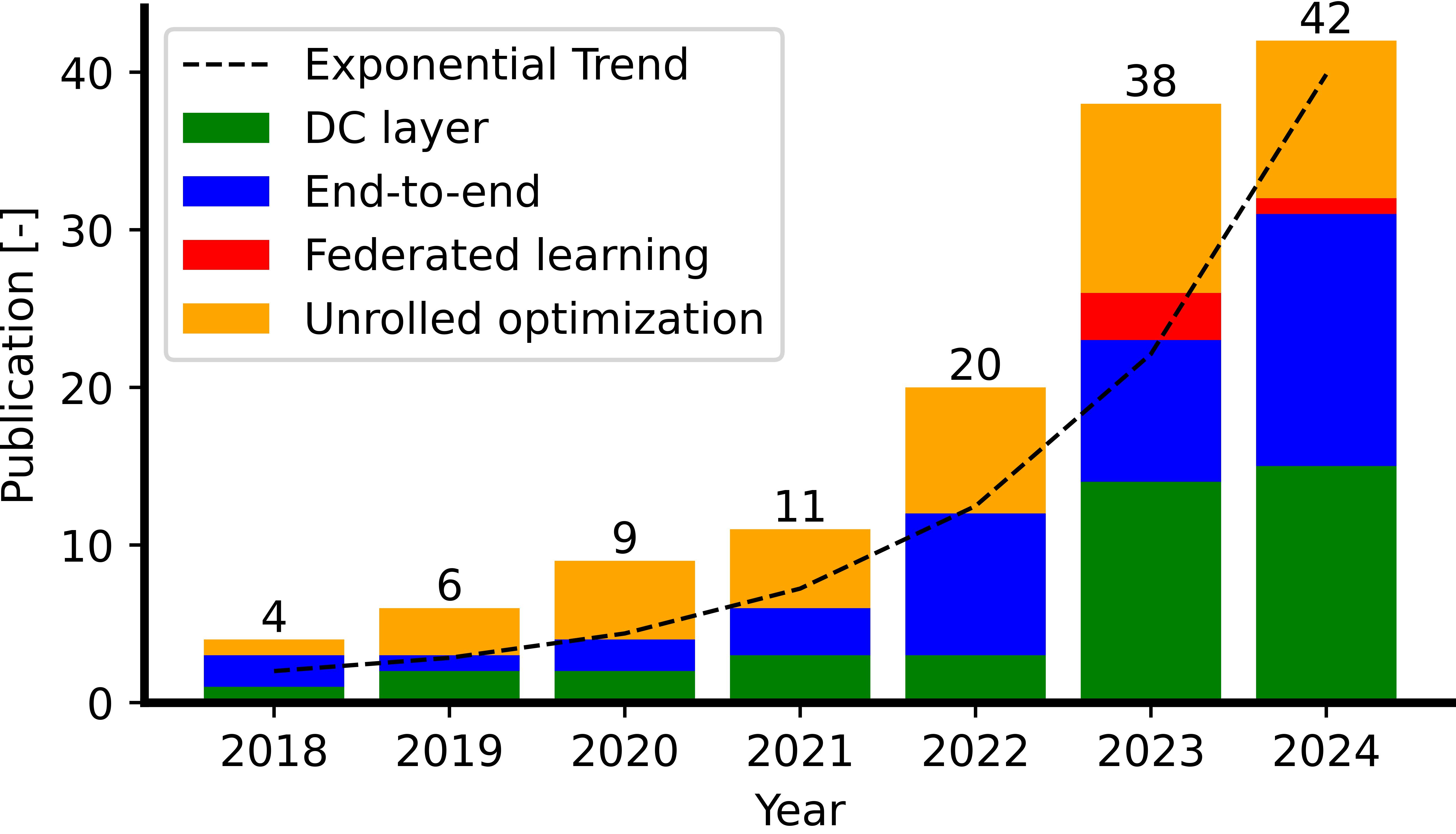}
	\caption{Overview of publications in DL-based MRI reconstruction models over time. The dashed line indicates the general trend plotted using $ 1+\exp(0.62 \times t) $ where t is defined in years.}
	\label{fig:trend_approaches_v01}
\end{figure}

Figure~\ref{fig:trend_approaches_v01} shows a stacked chart of the number of publications since 2018 by category. The total number of publications has grown exponentially in recent years and interest in the DC layer method continues to increase while interest in unrolling optimization and end-to-end approaches remains consistent over time.

\subsection{End-to-end models}

DL end-to-end models are specifically designed to tackle the MRI reconstruction problem without enforcing any data acquisition model. To achieve this, these models rely on a neural network to accurately predict fully sampled data from undersampled data~\cite{levac_accelerated_2023,mardani_deep_2018,yang_deep_2016}. Additionally, these models are trained using various regularization techniques that help address the ill-posed inverse problem. Table~\ref{tab:table3_reqularizations} lists common regularization techniques utilized in these models.

The end-to-end approach readily employs the same baseline models that are used for image-to-image translation, such as the U-net~\cite{hyun_deep_2018,xiang_ultra-fast_2018}, Swin transformers~\cite{huang_swin_2022} , and GANs~\cite{shitrit_accelerated_2017,zhao_swingan_2023}, making it easy to implement and allowing for the use of very deep and complex networks. However, they require a larger sample dataset than unrolling MRI reconstruction models and tend to predict images with synthetic data. Table~\ref{tab:end_to_end_papers_01} provides a list of selected references that used end-to-end DL models to solve DL-based MRI reconstruction algorithms.

\begin{table}[tbhp!]
	\caption{Overview of supervised end-to-end models to predict the fully sampled images.}
	\label{tab:end_to_end_papers_01}
	\resizebox{\textwidth}{!}{%
		\begin{tabular}{p{2cm} p{3cm} p{2cm} p{3cm} p{2cm} p{2.5cm} p{3cm}p{2cm}} 
			\hline

			DL Model      & Dataset                                                                & Region                                                          & Modality                                                               & sample size                                                 & R                                           & Ref.                         & Code\textsuperscript{1} \\ \hline
			
			GAN           & fastMRI\textsuperscript{2}~\cite{zbontar_fastmri_2018}                                        & Knee                                                            & PDFS,PD                                                                & 299300\textsuperscript{3}                                                     & 4,6                                         & \cite{narnhofer_inverse_2019}     & No    \\
			
			\rowcolor[HTML]{EFEFEF}GAN           & \begin{tabular}[c]{@{}l@{}}IXI\textsuperscript{4}\\ NAMIC\textsuperscript{5}\end{tabular}                  & Brain                                                           & \begin{tabular}[c]{@{}l@{}}T2,PD\\  T1,T2\end{tabular}                 & \begin{tabular}[c]{@{}l@{}}72\\ 20\end{tabular}             & 2,3,4                                       & \cite{lyu_multi-contrast_2020}         & No    \\
			
			ResNet        & Hammernik \textit{et al.}~\cite{hammernik_learning_2018}                                            & knee                                                            & PD                                                                     & 15                                                          & 5,7,9                                       & \cite{lu_pfista-sense-resnet_2020}           & Yes   \\
			
			\rowcolor[HTML]{EFEFEF}U-net         & Institutional\textsuperscript{6}                                                         & Knee                                                            & T1-map, T2-map                                                         & 10                                                          & 2,4,5,6                                     &  \cite{li_supermap_2023}             & No    \\
			
			GAN           & \begin{tabular}[c]{@{}l@{}}MICCAI 2013\textsuperscript{7}\\ ASC\textsuperscript{8} \\ fastMRI\end{tabular} & \begin{tabular}[c]{@{}l@{}}Brain\\ Cardiac \\ Knee\end{tabular} & \begin{tabular}[c]{@{}l@{}}T1\\ T1c \\ PD\end{tabular}                 & \begin{tabular}[c]{@{}l@{}}150\\ 100\\ 96\end{tabular}      & 2,3.3,10                                    &  \cite{gao_hierarchical_2023}            & No    \\

			\rowcolor[HTML]{EFEFEF}Y-net         & Own                                                                    & Brain                                                           & T1,T2                                                                  & 37                                                          & 2,3,4,5,6,8                                 & \cite{do_reconstruction_2020}                  & No    \\
			
			U-net         & Own                                                                    & Knee                                                            & T1                                                                     & 10                                                          & 3,4,5                                       & \cite{ayde_deep_2022}         & No    \\

			\rowcolor[HTML]{EFEFEF}SwinGAN       & \begin{tabular}[c]{@{}l@{}}IXI\\ MICCAI 2013\\ MRNet\end{tabular}      & \begin{tabular}[c]{@{}l@{}}Brain\\ Brain\\ Knee\end{tabular}    & \begin{tabular}[c]{@{}l@{}}T1\\ NS\textsuperscript{9}\\ NS\end{tabular}                  & \begin{tabular}[c]{@{}l@{}}573\textsuperscript{3}\\ 24809\textsuperscript{3}\\ 24809\textsuperscript{3}\end{tabular} & 20,30,50                                    & \cite{zhao_swingan_2023}         & Yes   \\

			Dense-Unet    & MICCAI 2016                                                            & Brain                                                           & T2                                                                     & 5                                                           & 2,4,8                                       &  \cite{xiang_ultra-fast_2018}       & No    \\

			\rowcolor[HTML]{EFEFEF}U-net         & Institutional                                                          & Cardiac                                                         & bSSFP,GRE                                                              & 80                                                          & NS                                          & \cite{wang_high-resolution_2024}             & No    \\

			GAN (Pix2pix) & \begin{tabular}[c]{@{}l@{}}HCP\\ Institutional\end{tabular}            & Brain                                                           & \begin{tabular}[c]{@{}l@{}}T1,T2\\  T2-map\end{tabular}                & \begin{tabular}[c]{@{}l@{}}20\\ NS\end{tabular}             & 8                                           & \cite{meng_accelerating_2021}           & No    \\
			
			GAN       & Institutional            & Cardiac\textsuperscript{10}                      & DTI                                    & 30           & 2.6               & \cite{liu_accelerated_2023}      & No    \\

			\rowcolor[HTML]{EFEFEF}DeepADC-Net   & Institutional                                                          & Animal study                                                    & DWI                                                                    & 183                                                         & 4,8                                         & \cite{li_learning_2024}          & Yes   \\

			NPB-REC       & fastMRI                                                                & Brain, Knee                                                     & NS                                                                     & 5847,1167                                                   & 4,8,12                                      &  \cite{khawaled_npb-rec_2024} & Yes   \\

			\rowcolor[HTML]{EFEFEF}SCU-Net       & Institutional                                                          & Brain                                                           & T1,T2,T2FLAIR                                                          & 180                                                         & 1.78,2.39,2.91, 3.33,3.64,4, 4.27,4.44,4.71 & \cite{jin_improving_2023}           & No    \\

			DuDReTLU-net  & Uk Biobank\textsuperscript{11}                                                           & Cardiac                                                         & NS                                                                     & 9032                                                        & 20                                          &  \cite{hong_dual-domain_2023}         & Yes   \\

			\rowcolor[HTML]{EFEFEF}MA-RECON      & fastMRI           & Brain, Knee           & T1c, T2FLAIR, PD                   & 1000\textsuperscript{3}, 34742\textsuperscript{3}      & 4,8                                         &  \cite{avidan_ma-recon_2024}   & Yes  \\
			
			GCESS  & \begin{tabular}[c]{l}fastMRI \\ Hammernik \textit{et al.}\end{tabular} & \begin{tabular}[c]{l}Brain \\ Knee\end{tabular} & \begin{tabular}[c]{l}T2 \\ PD \end{tabular} & \begin{tabular}[c]{l}45 \\ 20 \end{tabular} & 4,8,10&\cite{Ma2024} &Yes \\
			
			\rowcolor[HTML]{EFEFEF}FEFA & \begin{tabular}[c]{l}IXI \\ FastMRI\end{tabular} & \begin{tabular}[c]{l}Brain \\ Brain\end{tabular}& \begin{tabular}[c]{l}PD, T2 \\ T1, T2\end{tabular}& \begin{tabular}[c]{l}578 \\ 340\end{tabular} & 4, 8, 16 & \cite{10607849} & Yes\\
			
			DUN-SA & \begin{tabular}[c]{l}IXI \\ FastMRI \\Institutional \\BraTS 2018\end{tabular} & \begin{tabular}[c]{l} Brain \\ Brain \\Brain \\Brain \end{tabular} & \begin{tabular}[c]{l} T2, PD \\ T1, T2 \\ T1, T2 \\ T1, T2, T2FLAIR, T1c \end{tabular} & \begin{tabular}[c]{l} 570 \\ 340 \\34 \\ 210 \end{tabular} & 4, 8 &  \cite{ZHANG2025103331} & No \\
			
			\rowcolor[HTML]{EFEFEF} DiffGAN & \begin{tabular}[c]{l}IXI \\ BraTS 2013\\ MRNet \\fastMRI \end{tabular} & \begin{tabular}[c]{l}Brain \\ Brain \\ Knee \\ Knee \end{tabular} & \begin{tabular}[c]{l}T1 \\ ns \\ns \\ns \end{tabular} & \begin{tabular}[c]{l} 46,480~\textsuperscript{3} \\ 3200~\textsuperscript{3}\\ 14,700~\textsuperscript{3} \\  973 \end{tabular} & 4, 8 &	\cite{ZHAO2024108}&	 No \\

			CEST-PSnetwork & Institutional & CEST & Brain & 28 & 4,6,8,11,14 & 	\cite{https://doi.org/10.1002/mrm.30089} & Yes \\

			\rowcolor[HTML]{EFEFEF} ETER-net&  \begin{tabular}[c]{l}FastMRI\\ Calgary-Campinas~\cite{10.3389/fnins.2022.919186} \end{tabular}  & \begin{tabular}[c]{l} Brain, Knee \\ Brain  \end{tabular} & \begin{tabular}[c]{l} T2, PD, PDFS \\ T1  \end{tabular} & \begin{tabular}[c]{l} 379, 701, 701 \\ 67   \end{tabular} & 4 & \cite{OH2024108098} &  Yes \\

			Re-Con-GAN & Institutional & Lung & T1 & 48 & 3, 6, 10 & \cite{Xu_2024} & Yes\\

			\hline
			
			\multicolumn{8}{l}{\begin{tabular}[c]{@{}l@{}}
					
					1 Whether the original implementation is publicly available, \\ 
					
					2 \url{https://fastmri.med.nyu.edu/}, \\ 
					
					3 Number of image slices was specified, \\
					
					4 \url{https://brain-development.org/ixi-dataset/},\\ 
					
					5 \url{https://insight-journal.org/},\\ 
					
					6 The study used its Institutional dataset, \\ 
					
					7 MICCAI 2013 grand challenge dataset:  \\{\footnotesize \url{https://wiki.cancerimagingarchive.net/display/Public/NCI-MICCAI+2013+Grand+Challenges+in+Image+Segmentation}},\\ 
					
					8 Atrial Segmentation Challenge: \url{https://www.cardiacatlas.org/atriaseg2018-challenge/},\\ 
					
					9 Information not available from the publication, \\

					10 Ex-vivo study,\\ 
					
					11 \url{https://www.ukbiobank.ac.uk/},\\ 
					
					
			\end{tabular}}
		\end{tabular}%
	}
\end{table}

\subsection{unrolled model}
\subsubsection{unrolled optimization}

Unrolling in the DL-based MRI reconstruction context means transforming the iterative steps of an optimization algorithm into a deep neural network architecture, particularly ConvNets. Each iteration of the algorithm given in Equation~\ref{eq:equation01_cs} is mapped to a layer or a set of layers in the network. This approach allows the network to learn the parameters of the reconstruction process directly from training data, which can lead to more efficient and accurate reconstructions. The unrolled DL-based MRI reconstruction models have been shown to outperform end-to-end methods using a network with a smaller number of trainable parameters~\cite{aggarwal_j-modl_2020, geng_hfist-net_2023,liu_highly_2022,qiao_medl-net_2023,qu_radial_2024}.

However, the iterative nature of the unrolled optimization method may increase the computation time during both the training and inference steps, as the network’s weights are updated during training, and multiple iterative steps are performed during inference to ensure data consistency. Table~\ref{tab:unroll_optimization_all_list_v02} provides a list of references that used unrolled optimization models to solve DL-based MRI reconstruction algorithms.

\begin{table}[]
	\caption{Overview of supervised unrolled optimization to predict the fully sampled images.}
	\label{tab:unroll_optimization_all_list_v02}
	\resizebox{\textwidth}{!}{%
		
		\begin{tabular}{p{3cm} p{4cm} p{2cm} p{4cm} p{1.5cm} p{2.5cm} p{3cm}p{2cm}} 
			\hline
			
			DL Model            & Dataset                                                                                                                                        & Region                                                       & modality                                                                & sample size                                              & R                                             & Reference                & Code \\\hline
			ConvNet             & \begin{tabular}[c]{@{}l@{}}IXI\\ fastMRI\end{tabular}                                                                                          & \begin{tabular}[c]{@{}l@{}}Brain\\ Knee\end{tabular}         & \begin{tabular}[c]{@{}l@{}}T1,T2\\ PD\end{tabular}                      & \begin{tabular}[c]{@{}l@{}}50\\ 1025\end{tabular}        & 2.5, 3.3, 5                                   & \cite{liu_deep_2020}    & Yes  \\
			\rowcolor[HTML]{EFEFEF}ConvNet             & \begin{tabular}[c]{@{}l@{}}Institutional\\ fastMRI\end{tabular}                                                                                & \begin{tabular}[c]{@{}l@{}}Brain\\  Knee\end{tabular}        & \begin{tabular}[c]{@{}l@{}}T1 \\ PD, PDFS\end{tabular}                  & \begin{tabular}[c]{@{}l@{}}5\\ 60\end{tabular}           & 4,5,6.7                                       & \cite{guo_joint_2023}    & No   \\
			RG-Net              & Institutional                                                                                                                                  & Brain                                                        & T1$ \rho $                                                                      & 8                                                        & 17                                            & \cite{huang_densely_2017}     & No   \\
			\rowcolor[HTML]{EFEFEF}ConvNet             & Institutional                                                                                                                                  & Brain                                                        & SWI                                                                     & 117                                                      & 5,8                                           & \cite{duan_accelerating_2022}     & No   \\
			
			ConvNet             & HCP \cite{van_essen_wu-minn_2013}               & Brain                                                        & T1w,T2w                                                                 & 1200                                                     & NS\textsuperscript{1}                                           & \cite{zufiria_feature-based_2022}   & No   \\
			
			\rowcolor[HTML]{EFEFEF}ConvNet             & Institutional                                                                                                                                  & Cardiac, Abdominal                                           & NS                                                                      & 20,16                                                    & 3,4,5                                         & \cite{zhou_parallel_2019}      & No   \\
		
			RecurrentVarNet     & \begin{tabular}[c]{@{}l@{}}fastMRI\\ fastMRI \\ Calgary-Campinas\textsuperscript{2}\end{tabular}                                        & \begin{tabular}[c]{@{}l@{}}Brain\\ Knee\\ Brain\end{tabular} & \begin{tabular}[c]{@{}l@{}}T1c, T1, T2,T2-FLAIR\\ PD\\ T1w\end{tabular} & \begin{tabular}[c]{@{}l@{}}5846\\ 1172\\ 67\end{tabular} & 2,4,6                                         & \cite{YIASEMIS202433}  & No   \\

			\rowcolor[HTML]{EFEFEF}ConvNet             & \begin{tabular}[c]{@{}l@{}}Hammernik \textit{et al.}\\ Institutional\end{tabular}                                                               & \begin{tabular}[c]{@{}l@{}}Knee\\ Brain\end{tabular}         & \begin{tabular}[c]{@{}l@{}}PD\\ T1\end{tabular}                         & \begin{tabular}[c]{@{}l@{}}20\\ 8\end{tabular}           & 3,4,6                                         & \cite{hammernik_learning_2018} & No   \\
			
			UFLoss              & \begin{tabular}[c]{@{}l@{}}fastMRI\\ MRIdata\end{tabular}                                                                                      & \begin{tabular}[c]{@{}l@{}}Knee\\ Knee\end{tabular}          & \begin{tabular}[c]{@{}l@{}}PDFS, PD\\ PDFS, PD\end{tabular}             & \begin{tabular}[c]{l}380\\ 20\end{tabular}         & \begin{tabular}[c]{@{}l@{}}5\\ 8\end{tabular} & \cite{wang_high_2022} & Yes  \\
			
			\rowcolor[HTML]{EFEFEF}DFSN                & \begin{tabular}[c]{@{}l@{}}SRI24~\cite{rohlfing_sri24_2010} \\  MRBrainS13~\cite{mendrik_mrbrains_2015} \\ NeoBrainS12~\cite{isgum_evaluation_2015}\end{tabular} & Brain                                                        & \begin{tabular}[c]{@{}l@{}}PD,T1,T2\\ T1,T2-FLAIR \\ T1,T2\end{tabular} & \begin{tabular}[c]{@{}l@{}}36\\ 20\\ 175\end{tabular}    & 5, 10                                         & \cite{sun_deep_2019}       & No   \\
			
			ConvNet             & Institutional                                                                                                                                  & Brain                                                        & T1map                                                                   & 3                                                        & 8,12,18,36                                   & \cite{slavkova_untrained_2023}  & Yes  \\
			
			\rowcolor[HTML]{EFEFEF}JDL                 & Institutional                                                                                                                                  & Brain                                                        & T1,T2,PD,T2-FLAIR                                                        & 8                                                        & 4,8                                           & \cite{ryu_accelerated_2021} & Yes  \\
			pFISTA-DR           & Institutional                                                                                                                                  & Brain                                                        & T1,PD                                                                   & 200                                                      & 5,7,10                                        & \cite{qu_radial_2024}        & No   \\
			\rowcolor[HTML]{EFEFEF}U-net               & fastMRI                                                                                                                                        & Knee                                                         & PD, PDFS                                                                & 20                                                       & 4,6                                           & \cite{qiao_medl-net_2023}   & Yes  \\
		
			ConvNet             & \begin{tabular}[c]{@{}l@{}}fastMRI\\ Institutional\end{tabular}                                                                                & Brain                                                        & T1,T2,T2-FLAIR                                                          & \begin{tabular}[c]{@{}l@{}}120\\ 3\end{tabular}          & 1.8,2.5,3.5,4                                 & \cite{pramanik_adapting_2023}  & No   \\
			
			\rowcolor[HTML]{EFEFEF}ConvNet             & \begin{tabular}[c]{@{}l@{}}NAMIC\textsuperscript{3}\\ MRBrainS\textsuperscript{4}\end{tabular}                                                                                     & \begin{tabular}[c]{@{}l@{}}Brain\\ Brain\end{tabular}        & \begin{tabular}[c]{@{}l@{}}T1,T2\\ T2,T2-FLAIR\end{tabular}             & \begin{tabular}[c]{@{}l@{}}5\\ 7\end{tabular}            & 8                                             & \cite{liu_regularization_2021}    & No   \\
			Dictionary learning & Institutional                                     & Cardiac                                                      & CMR                                                                       & 19                                                       & 2,4,8                                         & \cite{kofler_deep_2023} & Yes  \\
			\rowcolor[HTML]{EFEFEF}PD-PCG-Net          & fastMRI                                                                                                                                        & Knee                                                         & PD, PDFS                                                                & 484489                                                   & 4                                             & \cite{kim_cascade_2022}    & No   \\
			RUN-UP              & Institutional                                                                                                                                  & Brain, Breast                                                & DTI, DWI                                                                & 14,6                                                     & 1004                                          & \cite{hu_run-up_2021}     & No   \\
			\rowcolor[HTML]{EFEFEF}ConvNet             & Institutional                                                                                                                                  & Cardiac                                                      & NS                                                                      & 22                                                       & 10,12,14                                      & \cite{sandino_accelerating_2021} & No   \\
			CEST-VN             & Institutional                                                                                                                                  & Brain                                                        & CEST                                                                    & 54                                                       & 3,4,5                                         & \cite{xu_accelerating_2024} & Yes  \\
			\rowcolor[HTML]{EFEFEF}Diffusion model     & GLIS-RT            & Brain              & T1c               & 230      & 1.25,1.66,2.5,5                 & \cite{safari_mri_2024}    & No   \\ 
			KIKI-net &  \begin{tabular}[c]{@{}l@{}}ADNI~\cite{https://doi.org/10.1002/jmri.21049}\\ Institutional\end{tabular}  & Brain & T2-FLAIR,T1c & 1500 image slices& 2,3,4 & \cite{https://doi.org/10.1002/mrm.27201} & No \\

			\rowcolor[HTML]{EFEFEF}Joint-ICNet & fastMRI & Brain & T2-FLAIR, T1c, T1w, T2w & 4,469 & 4,8 & \cite{9578412} & No \\
			
			E2E-VN & fastMRI & Brain, Knee & NS & NS & 4,6,8 & \cite{10.1007/978-3-030-59713-9_7} & No\\
			
			\rowcolor[HTML]{EFEFEF} vSHARP &CMR$ \times $Recon(2023) & Cardiac & CMR, T1w, T2w & 314\textsuperscript{5}, 358\textsuperscript{6} & 4, 8, 10 & \cite{10.1007/978-3-031-52448-6_45} & No \\
			
			vSHARP & \begin{tabular}[c]{l}Calgary-Campinas\\ fastMRI\\ CMR$ \times $Recon(2023)\end{tabular}    & \begin{tabular}[c]{l}Brain\\ Prostate\\ Cardiac\end{tabular} &  \begin{tabular}[c]{l} T1w \\ T2w \\ Cine MRI \end{tabular} &  \begin{tabular}[c]{l} 67 \\ 312 \\ 473 \end{tabular} & 4, 8, 16 & \cite{YIASEMIS2025110266} & No \\
			\rowcolor[HTML]{EFEFEF}GrappaNet &  fastMRI & Knee & PD, PDFS & 1,594 & 4, 8 & \cite{9157643} & No \\
			MoDL & Institutional & Brain &T2w & 5& 6, 10 & \cite{aggarwal_modl_2018} & Yes\\	
			\rowcolor[HTML]{EFEFEF}VS-Net & Hammernik \textit{et al.}& Knee & PD, PDFS, T2w, T2FS & 20 & 4, 6 & \cite{10.1007/978-3-030-32251-9_78} & Yes \\

			VSNet & Institutional & Cardiac & bSSFP, GRE & 61 & 2.3, 2.7, 4, 5 & \cite{https://doi.org/10.1002/jmri.29295} & No\\
			 
			\rowcolor[HTML]{EFEFEF}CineVN & \begin{tabular}[c]{l}Institutional \\ OCMR\end{tabular}  & Cardiac & Cardiac & \begin{tabular}[c]{l}64 \\ 165\end{tabular} &\begin{tabular}[c]{l}2,8,16 \\ 8,12,16,20\end{tabular} &\cite{https://doi.org/10.1002/mrm.30260} & Yes \\
			
			Distill & \begin{tabular}[c]{l} Institutional \\ fastMRI \end{tabular}& \begin{tabular}[c]{l} Brain\\ Knee \end{tabular}& \begin{tabular}[c]{l} T1 \\ PD, PDFS \end{tabular}& \begin{tabular}[c]{l} 17 \\ 232 \end{tabular} & 10 &\cite{10433659} & Yes\\
			
			\rowcolor[HTML]{EFEFEF} ConvLSTM &  AMRG\textsuperscript{7} & Cardiac & Cardiac & 1200 slices & 8,10,12& \cite{https://doi.org/10.1002/mp.17501} & No\\
			
			RELAX-MORE & Institutional & Brain, Knee & T1-map & 20 & 3 & \cite{https://doi.org/10.1002/mrm.30045} & No\\

			\hline
			\multicolumn{8}{l}{\begin{tabular}[c]{@{}l@{}}1 Not specified\\ 
					2 Calgary-Campinas dataset was released as part of the Multi-Coil MRI Reconstruction Challenge\\
					3 \url{http://hdl.handle.net/1926/1687}\\ 4 \url{https://mrbrains18.isi.uu.nl/} \\
					5 Sample size of Cine MRI data \\
					6 Sample size of multi-contrast data T1w and T2w images\\
				    7 \url{https://www.cardiacatlas.org/amrg-cardiac-atlas/}                                  \end{tabular}}                                                                                                                                                                                                                                                
		\end{tabular}%
	}
\end{table}

\subsubsection{Data consistency layer}

The more popular approach is to train the unrolled models similarly to the end-to-end models as follows:

\begin{equation}
	\centering
	x_{\hat{f}_{\psi}} = \arg\min \| x - f_{\psi}(x_{\Omega}) \|_1 + \lambda \underbrace{\| y_{\Omega} - E_{\Omega} x \|_2^2}_{\text{Data consistency}}
	\label{eq:dclayer_}
\end{equation}
where $ \lambda $ is a penalty term that balances the DC and prior term. $ f_\psi $ is a DL model that maps the undersampled input images $ x_\Omega $ to reconstruct fully sample images $ x $. The DL reconstruction and data consistency operate on the image domain and \textit{k}-space domain, respectively. Although the DL part is trained without incorporating a priori information, the second term discourages the DL first part from updating the \textit{k}-space parts that were not sampled~\cite{schlemper_deep_2017}. The closed form for (\ref{eq:dclayer_}) is as follows~\cite{qin_convolutional_2018}:

\begin{equation}
	\centering
	\hat{X} = \begin{cases}
		\hat{X}_{{f}_{\psi}}(k) & \text{if } k \notin \Omega \\
		\frac{\hat{X}_{\hat{f}_{\psi}}(k) + \lambda_0 X_{\Omega}(k)}{\lambda_0 + 1} & \text{Otherwise}
	\end{cases}
	\label{eq:closedform_}
\end{equation}
where $ \lambda_0 $ is related to the image noise level and in the limit $ \lambda_0 \to \infty $ for noise-free images. Additionally,  \textit{k} refers to the \textit{k}-space lines and $ \hat{X} $ and $ X_\Omega $ are the Fourier transform of $ \hat{x} $ and $ x_\Omega $, respectively. This closed form is a computational layer called the DC layer at the end of DL models. The DC layer is a crucial part of the MRI reconstruction DL models, playing an important role in reconstructing images~\cite{cheng_learning_2021,korkmaz_self_supervised_2023}. The DC layer allows for a flexible design of the DL model when it is added to U-net~\cite{murugesan_deep_2021}, transformers~\cite{wu_deep_2023}, stable diffusion model~\cite{cao_high-frequency_2024}, and so on. Table~\ref{tab:dc_layer_all_papers} summarizes the DL-based MRI reconstruction models trained under the DC layer framework.

\begin{table}[tbh!]
	\caption{Overview of supervised models that used the data consistency to predict the fully sampled images.}
	\label{tab:dc_layer_all_papers}
	\resizebox{\textwidth}{!}{%
		
		\begin{tabular}{p{3cm} p{4cm} p{2cm} p{6cm} p{1.5cm} p{2.5cm} p{3cm}p{2cm}} 
			\hline
			
			DL Model       & Dataset                                                                                  & Region                                                                              & Modality                                                                                              & sample size                                                    & R              & Ref.                      & Code \\\hline
			KTMR           & Institutional                                                                            & Brain                                                                               & T1c, MRA                                                                                              & 17                                                             & 2,2.5,3.3,5,10 & \cite{wu_deep_2023}  & Yes  \\
			\rowcolor[HTML]{EFEFEF}DCT-net        & \begin{tabular}[c]{@{}l@{}}fastMRI\\ Calgary\end{tabular}                                & \begin{tabular}[c]{@{}l@{}}Brain\\ Knee\end{tabular}                                & \begin{tabular}[c]{@{}l@{}}T1\\ PD\end{tabular}                                                       & \begin{tabular}[c]{@{}l@{}}973\\ 25\end{tabular}               & 4,8            & \cite{wang_dct-net_2024}    & No   \\
			CTFNet         & Institutional                                                                            & Cardiac                                                                             & bSSFP                                                                                                 & 48                                                             & 8,16,24        & \cite{qin_complementary_2021}        & Yes  \\
			\rowcolor[HTML]{EFEFEF}EDAEPRec       & Institutional                                                                            & Brain                                                                               & T2                                                                                                    & 7                                                              & 3.3,4,5,6.7,10 & \cite{liu_highly_2020}   & No   \\
			GFN            & Institutional                                                                            & Brain                                                                               & T1w,T2FLAIR,TOF                                                                                  & 30,50,80                                                       & 3.3,5,10       & \cite{dai_deep_2023}        & No   \\
			\rowcolor[HTML]{EFEFEF}PC-RNN         & fastMRI                                                                                  & \begin{tabular}[c]{@{}l@{}}Knee\\ Brain\end{tabular}                                & \begin{tabular}[c]{@{}l@{}}PD, PDFS\\ T1,T2\end{tabular}                                              & \begin{tabular}[c]{@{}l@{}}973\\ 4469\end{tabular}             & 4,6            & \cite{chen_pyramid_2022} & Yes  \\
			ResNet Unet    & Institutional                                                                            & Knee                                                                                & T1                                                                                                    & 360                                                            & 6              & \cite{wu_self-attention_2019}         & No   \\
			\rowcolor[HTML]{EFEFEF}CNF            & fastMRI                                                                                  & Knee, Brain                                                                         & PD, T2                                                                                                & 20,8                                                           & 4              & \cite{wen_conditional_2023}        & Yes  \\
			stDLNN         & Institutional                                                                            & Abdomen                                                                             & GRE                                                                                                   & 8                                                              & 4,8,16,25      & \cite{wang_parallel_2023}    & Yes  \\
			\rowcolor[HTML]{EFEFEF}MODEST         & Institutional                                                                            & Cardiac                                                                             &                                                                                                       & 28                                                             & 3.7,7.4,14.8   & \cite{terpstra_accelerated_2023}   & Yes  \\
			DeepcomplexMRI & \begin{tabular}[c]{@{}l@{}}Institutional\\ Hammernik \textit{et al.}\end{tabular}     & \begin{tabular}[c]{@{}l@{}}Brain\\ Knee\end{tabular}                                & \begin{tabular}[c]{@{}l@{}}T1,T2,PD \\ PD\end{tabular}                                                & \begin{tabular}[c]{@{}l@{}}22\\ 20\end{tabular}                & 4,5,10         & \cite{wang_deepcomplexmri_2020} & No   \\
			\rowcolor[HTML]{EFEFEF}AdaDiff        & \begin{tabular}[c]{@{}l@{}}fastMRI\\ IXI\end{tabular}                                    & Brain                                                                               & \begin{tabular}[c]{@{}l@{}}T1, T2, PD\\ T1,T2,T2-FLAIR\end{tabular}                                   & \begin{tabular}[c]{@{}l@{}}66\\ 420\end{tabular}               & 4,8            & \cite{gungor_adaptive_2023}   & Yes  \\
			McSTRA         & fastMRI                                                                                  & Knee                                                                                & PD, PDFS                                                                                              & 584, 588                                                       & 4,6,8,10,12    & \cite{ekanayake_mcstra_2024}  & No   \\
			\rowcolor[HTML]{EFEFEF}DC-RSN         & \begin{tabular}[c]{@{}l@{}}ACDC~\cite{bernard_deep_2018}\\ Kirby\\ Calgary\\ MRBrain\\ Hammernick\end{tabular} & \begin{tabular}[c]{@{}l@{}}Cardiac\\ Brain\\ Brain \\ Brain,\\ Knee\end{tabular} & \begin{tabular}[c]{@{}l@{}}Cardiac, \\ T1-MPRAGE\\ T1\\ T2-FLAIR\\ PDF, PD, T1, T2, T2FS\end{tabular} & \begin{tabular}[c]{@{}l@{}}200\\ 42\\ 45\\ 7\\ 25\end{tabular} & 4, 5           & \cite{murugesan_deep_2021}  & No  \\

			NF-cMRI & El‐Rewaidy \textit{et al.}~\cite{el2021multi} & Cardiac  & bSSFP cine& 108 & 13, 17, 24 & \cite{CATALAN2025109467} & Yes \\
			
			\rowcolor[HTML]{EFEFEF} CST-Net & OCMR~\cite{chen2020ocmrv10openaccessmulticoilkspace} & Cardiac&Cardiac &102& 6, 12&\cite{WANG2024109133} & Yes\\
			
			DIRECTION & fastMRI & Knee & ns & 1172 & 4, 8 & \cite{SUN2024157} & No \\

			\rowcolor[HTML]{EFEFEF} SPICER & \begin{tabular}[c]{l} Institutional \\ fastMRI \end{tabular} & \begin{tabular}[c]{l} Brain \\ Brain \end{tabular} & \begin{tabular}[c]{l} T1 \\ T2 \end{tabular} & \begin{tabular}[c]{l} 20 \\ 165 \end{tabular} & 4, 6, 8 & \cite{https://doi.org/10.1002/mrm.30121} & Yes\\
			
			MLMFNet & \begin{tabular}[c]{l} BraTS 2019 \\ fastMRI \end{tabular} & \begin{tabular}[c]{l} Brain \\ Knee \end{tabular} & \begin{tabular}[c]{l} T1, T2 \\ PD, PDFS \end{tabular} & \begin{tabular}[c]{l} 212 \\ 243 \end{tabular} & 4, 8 & \cite{ZHOU2024246} & No \\

			\rowcolor[HTML]{EFEFEF}MCCA & \begin{tabular}[c]{l}fastMRI \\ Institutional \end{tabular} & \begin{tabular}[c]{l}Knee \\ Brain \end{tabular}& \begin{tabular}[c]{l}PD, PDFS \\ T1, T2 \end{tabular}& \begin{tabular}[c]{l}229 \\ 188 \end{tabular} & 4, 8, 12 &\cite{10376268} & Yes \\
			
			DiffINR & \begin{tabular}[c]{l}fastMRI \\ Institutional \end{tabular} &\begin{tabular}[c]{l}Knee, Brain \\ Knee \end{tabular} &\begin{tabular}[c]{l}PD, PDFS, T2FLAIR \\ T1 \end{tabular} &\begin{tabular}[c]{l}204 \\ 1\textsuperscript{1} \end{tabular} & 8,12,15,18 & \cite{CHU2025103398} & No \\
			
			\rowcolor[HTML]{EFEFEF}SCAMPI & fastMRI &Brain & T1, T2, T2FLAIR & 200 & 3, 5, 8& \cite{https://doi.org/10.1002/mrm.30114} & No \\

			\hline
			\multicolumn{8}{l}{\begin{tabular}[c]{@{}l@{}}
					
					1 It was used to evaluate model's generalization, \\

			\end{tabular}}
		\end{tabular}%
	}
\end{table}

\subsection{Federated learning}\label{sub:federatedLearning}
Federated learning (FL) is a promising framework that enables the collaborative training of learning-based models across multiple institutions without the need for sharing local private data~\cite{yang_federated_2019}. The objective of FL models is to learn a global model by taking the average of local models~\cite{mcmahan_communication-efficient_2017} or by ensuring the proximity of local models to the global model~\cite{li_federated_2020}. When applied to MRI image reconstruction, the FL offers unique advantages tailored to the specific challenges and requirements as follows:

\begin{itemize}
	\item MR images often contain sensitive patient information that needs to be protected. FL enables MRI models to be trained directly on the devices where the images are acquired without the need to transmit patient data to a centralized location. This decentralization of data ensures that patient privacy and confidentiality are maintained.
	\item MRI machines can vary in their hardware specifications and imaging protocols, which can lead to challenges in standardizing image reconstruction algorithms. However, FL accommodates this heterogeneity by allowing models to be trained collaboratively across different types of MRI machines, ensuring that the reconstruction algorithms are robust and adaptable to various configurations.
	
\end{itemize}

It is worth noting that FL models are predominantly supervised and have been developed under the end-to-end and unrolled model frameworks, which have shown promising results in various applications. Table~\ref{tab:fl_list_all_papers} summarizes the FL framework for the DL-based MRI models.

\begin{table}[tbh!]
	\caption{Overview of supervised models that were trained with federated learning.}
	\label{tab:fl_list_all_papers}
	\resizebox{\textwidth}{!}{%
		
		\begin{tabular}{p{2cm} p{3cm} p{5cm} p{1.5cm} p{2cm}p{1cm}} 
			\hline
			
			DL Model        & Training framework & Dataset                                                                                                          & sample size                                                     & Ref.                       & Code \\\hline
			FL-MRCM         & End-to-End         & \begin{tabular}[c]{@{}l@{}}fastMRI, \\ HPKS\cite{jiang_identifying_2019}\\ IXI,\\  BraTS~\cite{menze_multimodal_2014}\end{tabular} & \begin{tabular}[c]{@{}l@{}}3443\\ 144\\ 434\\ 494\end{tabular}  & \cite{guo_multi-institutional_2021}& Yes  \\
			\rowcolor[HTML]{EFEFEF}FedMRI          & End-to-End         & \begin{tabular}[c]{@{}l@{}}fastMRI\\ BraTS\\ SMS~\cite{feng_multi-contrast_2021}\\ uMR~\cite{feng_multi-contrast_2021}\end{tabular}      & \begin{tabular}[c]{@{}l@{}}2134\\ 385\\ 155\\ 50\end{tabular}   & \cite{feng_specificity-preserving_2022} & Yes  \\
			ACM-FedMRI      & DC guided          & \begin{tabular}[c]{@{}l@{}}fastMRI\\ BraTS\\ IXI\\ Institutional\end{tabular}                                    & \begin{tabular}[c]{@{}l@{}}3443\\ 494 \\ 526\\ 150\end{tabular} & \cite{lyu_adaptive_2023}      & No   \\
			\rowcolor[HTML]{EFEFEF}FedGIMP         & DC guided          & \begin{tabular}[c]{@{}l@{}}fastMRI \\ BraTS\\ IXI\\ Institutional\end{tabular}                                   & \begin{tabular}[c]{@{}l@{}}51 \\ 55\\ 55\\ 10\end{tabular}      & \cite{elmas_federated_2022}       & Yes  \\
			Unrolled FedLrn & Unrolled             & fastMRI, MRIdata                                                                                                 & NS                                                              & \cite{levac_federated_2023} & Yes  \\
			
			\rowcolor[HTML]{EFEFEF}GAutoMRI & Unrolled &\begin{tabular}[c]{l}fastMRI \\ Calgary-Campinas\\ Aggarwal~\textit{et al.}~\cite{aggarwal_modl_2018}\\ Institutional \end{tabular} & \begin{tabular}[c]{l}570 \\ 35\\ 542 slices\\ 22 \end{tabular} &\cite{10606250} & Yes\\
			
			\hline

		\end{tabular}%
	}
\end{table}

\begin{table}[tbh!]
	\caption{Overview of self-supervised models to predict the fully sampled images.}
	\label{tab:self_supervised_all_papers}
	\resizebox{\textwidth}{!}{%
		
		\begin{tabular}{p{3cm} p{4cm} p{2cm} p{4cm} p{1.5cm} p{2.5cm} p{2cm}p{1cm}} 
			\hline
			
			DL Model        & Dataset                                                                & Region                                                       & modality                                                           & sample size                                             & R     & Ref.                   & Code \\\hline
			DC-SiamNet      & \begin{tabular}[c]{@{}l@{}}IXI\\ MRINet\end{tabular}                   & \begin{tabular}[c]{@{}l@{}}Brain\\ Knee\end{tabular}         & \begin{tabular}[c]{@{}l@{}}T1, T2, PD \\ PD\end{tabular}           & \begin{tabular}[c]{@{}l@{}}473\\ 1250\end{tabular}      & 4, 5  & \cite{yan_dc-siamnet_2023} & No   \\
			\rowcolor[HTML]{EFEFEF}Noise2Recon     & \begin{tabular}[c]{@{}l@{}}MRIdata\\ fastMRI\end{tabular}              & \begin{tabular}[c]{@{}l@{}}Knee\\ Brain\end{tabular}         & \begin{tabular}[c]{@{}l@{}}PDFS \\ T2,T2-FLAIR,T1,T1c\end{tabular} & \begin{tabular}[c]{@{}l@{}}19\\ 603\end{tabular}        & 12,16 & \cite{desai_noise2recon_2023}   & Yes  \\
			DDSS            & HCP                                                                    & Brain                                                        & T1                                                                 & 505                                                     & 2,4   & \cite{zhou_dual-domain_2022}    & No   \\
			\rowcolor[HTML]{EFEFEF}SSDU            & \begin{tabular}[c]{@{}l@{}}Institutional\\ fastMRI\end{tabular} & \begin{tabular}[c]{@{}l@{}}Brain\\ Knee\\ Brain\end{tabular} & \begin{tabular}[c]{@{}l@{}}T1MPRAGE\\ PD,PDFS\\ T2\end{tabular}    & \begin{tabular}[c]{@{}l@{}}19\\ 35,25\\ 10\end{tabular} & 4,6,8 & \cite{yaman_self-supervised_2020}  & Yes  \\
			Multi-mask SSDU &       \begin{tabular}[c]{@{}l@{}}Institutional \\ MRIDdata\end{tabular}                               & \begin{tabular}[c]{@{}l@{}}Brain \\ Knee\end{tabular}        & \begin{tabular}[c]{@{}l@{}}T1MPRAGE \\ PD\end{tabular}             & \begin{tabular}[c]{@{}l@{}}9 \\ 20\end{tabular}         & 8,12  & \cite{yaman_multi_mask_2022}   & No   \\
			\rowcolor[HTML]{EFEFEF}RELAX           & Institutional                                                          & Brain, Knee                                                  & T1 and T2 maps                                                     & 21                                                      & 5     &\cite{liu_magnetic_2021}  & No   \\
			Joint MAPLE     & Institutional                                                          & Brain                                                        & T1 and T2* maps                                                    & 2                                                       & 16,25 & \cite{heydari_joint_2024} & Yes \\\hline

		\end{tabular}}
\end{table}

\subsection{Assessment}

The previous sections covered three major DL-based MRI reconstruction training approaches: end-to-end, unrolled optimization, and the DC layer. While these methods can be applied to both supervised and self-supervised frameworks, the unrolled optimization and DC layer approaches are typically used for self-supervised training (see Table~\ref{tab:self_supervised_all_papers}). A summary of the advantages and disadvantages of each approach is provided in Table~\ref{tab:assessment_lists}.

\begin{table}[tbh!]
	\caption{A list of the pros and cons of each training framework is provided.}
	\label{tab:assessment_lists}
	\resizebox{\textwidth}{!}{%
		\begin{tabular}{p{3cm}  p{9cm} p{9cm}} 
			\hline
			DL method           & Pros                                                                                                                                                                                                                                      & Cons                                                                                                                                                                                                                           \\\hline
			End-to-end          & \begin{tabular}[c]{@{}p{9cm}@{}}\raggedright 1. Free from MRI data acquisition method.\\ 2. Easy-to-employ models proposed for image synthesis and segmentation.\end{tabular}                                                                         & \begin{tabular}[c]{@{}p{9cm}@{}}\raggedright 1. Requires a big dataset.\\ 2. Likely to add unwanted image structure, specifically for a higher acceleration rate.\\ 3. Limited generalization to out-of-distribution data.\end{tabular} \\
			unrolled optimization & \begin{tabular}[c]{@{}p{9cm}@{}}\raggedright 1. Requires smaller datasets compared with the end-to-end method.\\ 2. More likely to generalize better.\\ 3. Performs well using a ConvNet with a small number of trainable parameters.\end{tabular} & \begin{tabular}[c]{@{}p{9cm}@{}}\raggedright 1. Iterative method.\\ 2. Requires image acquisition knowledge.\\ 3. Requires under-sampling pattern in inference time.\end{tabular}                                                       \\
			DC Layer            & \begin{tabular}[c]{@{}p{9cm}@{}}\raggedright 1. Requires a smaller dataset than the end-to-end method.\\ 2. Provides closed-form equation.\\ 3. Easy to implement.\end{tabular}                                                                   & \begin{tabular}[c]{@{}p{9cm}@{}}\raggedright 1. Requires under-sampling pattern in inference time.\end{tabular}                                                                                                                                                                     
			\ \\\hline
			
		\end{tabular}%
	}
\end{table}

\section{Benchmark Competitions in MRI Reconstruction}

Large-scale competitions are instrumental in driving the progress of DL methods for MRI reconstruction. By uniting diverse datasets, standardizing evaluation protocols, and establishing clear metrics for comparison, these events create an environment that encourages rigorous benchmarking of innovative architectures and algorithms. This not only accelerates methodological development but also enhances transparency in the field, ultimately benefiting researchers and patients.

A notable initiative is the Multi-Coil MRI Reconstruction Challenge~\cite{10.3389/fnins.2022.919186}~\footnote{\url{https://github.com/rmsouza01/MC-MRI-Rec}}, which aims to evaluate the generalizability of reconstruction models across different coil configurations. This challenge focuses on raw multi-coil 3D \textit{k}-space data of T1w brain images from 167 healthy volunteers. The datasets were obtained from 12-channel and 32-channel receiver coils, comprising 70\% and 30\% of the data, respectively. Participants could thoroughly evaluate algorithm performance across varied coil arrays. This carefully curated dataset highlights the strengths and weaknesses in coil-sensitivity estimation, artifact correction, and network generalization.

The fastMRI Challenge~\cite{9420272}, initiated by Facebook AI Research~\footnote{\url{https://github.com/facebookresearch/fastMRI}} in collaboration with NYU Langone Health~\footnote{\url{https://fastmri.med.nyu.edu/}}, presents a significant large-scale benchmark focused on raw multi-coil, primarily 2D \textit{k}-space data, with the exception of the breast dataset, which is acquired in 3D. This challenge encompasses a variety of tasks, ranging from reconstructions to more extensive clinical evaluations. Its publicly accessible dataset spans several anatomies, including brain (6,970 samples), knee (1,500 samples), breast (300 samples)~\cite{solomon2024fastmribreastpubliclyavailable}~\footnote{\url{https://github.com/eddysolo/demo_dce_recon}} and prostate (312 samples)~\cite{Tibrewala2024}~\footnote{\url{https://github.com/cai2r/fastMRI_prostate}} MRI images. This rich resource empowers researchers to test their techniques under standardized conditions and effectively identify the boundaries of generalization in their work.

Beyond simple reconstruction tasks, the K2S Challenge~\cite{bioengineering10020267}~\footnote{\url{https://k2s.grand-challenge.org/}} expands its focus by integrating undersampled multi-coil 3D \textit{k}-space acquisition from 300 patients with automated segmentation of knee tissue. This approach emphasizes the necessity of evaluating reconstructions for comprehensive performance. If undersampled data can be reconstructed accurately but fails during advanced post-processing, its practical application may be limited. K2S provides a cohesive platform for assessing these interconnected processes, drawing attention to the relationship between reconstruction quality and subsequent clinical workflows.

CMR$ \times $Recon~\cite{wang2023cmrxreconopencardiacmri,wang2024cmrxreconopencardiacmri_v2}~\footnote{\url{https://github.com/CmrxRecon/CMRxRecon/}}$ ^, $\footnote{\url{https://github.com/CmrxRecon/CMRxRecon2024/}} tackles critical challenges in cardiac MRI for 2023 and 2024, focusing on enhancements in reconstruction fidelity, motion compensation, and temporal resolution. These challenges provide multi-coil and single-coil raw \textit{k}-space data from 300 subjects, alongside manual segmentation of the myocardium and chambers. These challenges fill an essential gap, as cardiac imaging demands unique considerations due to distinct cardiac motion and rapid hemodynamic shifts. By rigorously testing DL-based reconstruction methods in these dynamic environments, the field has the potential to propel algorithms toward reliable performance in real-time or near-real-time clinical scenarios. This could ultimately lead to significant improvements in patient care and diagnostic accuracy.

These competitions collectively highlight the importance of open and well-documented benchmarking. They bring together communities of researchers, stimulate the exploration of new ideas, and help ensure that the latest methodological advances are both comparable and clinically relevant. As DL-based MRI reconstruction continues to evolve, ongoing and future challenges will play a crucial role in shaping a more reproducible, transparent, and effective environment for rapid MRI acquisition and analysis.

\section{Evaluation metrics}
\subsection{Quantitative metrics}\label{subsec:quantiative_metrics}

When testing MRI reconstruction models, the quality of reconstructed images is quantitatively assessed by comparing them with the ground truth. Retrospective under-sampling of the \textit{k}-space allows the reporting of indices to quantify the quality of the reconstructed images.

\begin{figure}[tbh!]
	\centering
	\includegraphics[width=\textwidth]{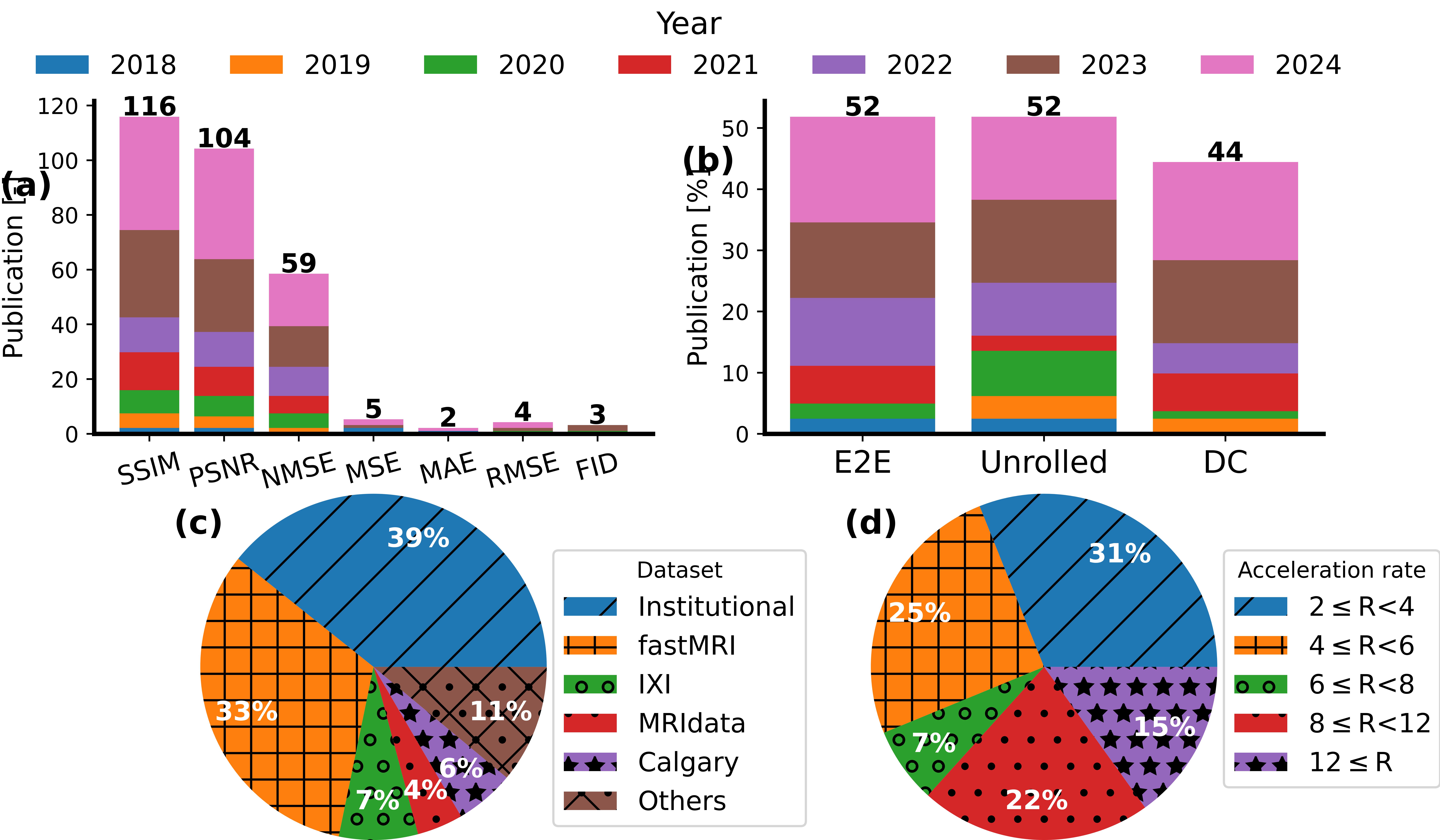}
	\caption{This graph illustrates (a) The metric used in different studies, (b) the training method used in different studies, (c) the dataset used for training used in the studies, and (d) the acceleration rate of the under-sampling. Abbreviations: SSIM: structural similarity index; PSNR: peak signal-to-noise ratio; NMSE: normalized mean square error; NRMSE: normalized root mean square error; MSE: mean square error; MAE: mean absolute error; RMSE: root mean square error; and FID: Fréchet inception distance.}
	\label{fig:metrics_all}
\end{figure}

Most studies quantitatively compare predicted images with ground truth images. As indicated in Figure~\ref{fig:metrics_all}a, most of these studies use the structural similarity (SSIM)~\cite{1284395} index and peak signal-to-noise ratio (PSNR) to compare reconstructed image with ground truth image. SSIM ranges between -1 and 1, with the best similarity achieved by an SSIM equal to one. A higher PSNR indicates a better reconstruction. The logarithmic operator quantifies image quality that closely aligns with human perception~\cite{safari_medfusiongan_2023}.

The normalized mean square error (NMSE) has become more popular since 2022 to quantify the quality of reconstructed images. Smaller NMSE values indicate better image reconstruction. Nonetheless, it might favor image smoothness rather than sharpness.

However, other metrics such as root mean square error, mean square error, mean absolute error, and Fréchet inception distance are rarely used, especially after a recommendation made in 2018 by Zbontar, \textit{et al.}~\cite{zbontar_fastmri_2018}(see Figure~\ref{fig:metrics_all}).

Regarding the training methods, the unrolled models, including the unrolled optimization and DC layer, are the most commonly used, with a growing use of the DC layer since 2020, which is expected to continue this trend in the future. More details about these trends are illustrated in Figure~\ref{fig:metrics_all}b.

Our systematic review found that around 39\%  of studies used institutional datasets, while fastMRI was the most frequently used public dataset. The majority of private datasets use single-coil images compared to the latter’s raw multi-coil raw 2D \textit{k}-space data. The fastMRI dataset consists of three imaging regions: the brain, pelvis, prostate, and knee regions. Less commonly used datasets include IXI, Calgary, and MRIdata, with only around 7\%, 6\%, and 4\% usage rates, respectively (see Figure~\ref{fig:metrics_all}c).

In addition, most studies reviewed by this study tested their proposed model using acceleration factor (R) $ 2 \le R < 6 $. The least simulated acceleration factor was $ R \ge 12 $ and $ 6 \le R < 8 $ with 14.6\% and 7.3\% of usage, respectively. The percentage of acceleration usage is summarized in Figure~\ref{fig:metrics_all}d.

\subsection{Clinical evaluations}

The metrics presented in Section~\ref{subsec:quantiative_metrics} quantify the quality of image reconstruction, but their results may not directly correlate with clinical outcomes. Several studies have been conducted to evaluate the clinical significance of MRI reconstruction using DL models. For example, a study found that DL-based MRI reconstruction models and fully sampled MRI images showed no significant differences (\textit{p}-values $> 0.05$) in the organ-based image quality of the liver, pancreas, spleen, and kidneys, number and measured diameter of the detected lesions while reducing the imaging time by more than 85\%~\cite{herrmann_comprehensive_2023}.

Similarly, another study showed that brain MRI images accelerated up to $ 4\times $ and $ 14\times $ had sufficient image quality for diagnostic and screening purposes, respectively~\cite{radmanesh_exploring_2022}. A third study found that there was no statistical significance (\textit{p}-value = 0.521) between the DL-based T2-FLAIR MRI image and standard T1c MRI images in the assessment of inflammatory knee synovitis~\cite{feuerriegel_inflammatory_2023}. These studies are consistent with another conducted to compare the diagnostic performance of DL MRI and standard MRI images in detecting knee abnormalities~\cite{johnson_deep_2023}.

A recent study found that there is no significant difference in the overall quality of MRI images generated by DL and standard fully sampled images for various MRI sequences, including T2 and diffusion-weighted imaging, for patients with prostate cancer. The study also revealed that DL MRI and standard MR images identified a similar number of Prostate Imaging Reporting and Data System $ \ge $ 3. However, the imaging time was significantly reduced by about 3.7-fold with the use of DL MRI. This study’s findings suggest that DL MRI can be a viable alternative to standard MRI imaging for prostate cancer patients, as it can produce similar quality images in a significantly shorter acquisition time~\cite{johnson_deep_2022}.

\section{Discussion}

This systematic review investigates the contemporary development of using DL for MRI acceleration to enhance reconstruction efficiency and image quality. In this work, we characterized the potentials and limitations of individual methods using DL or CS for MRI reconstruction and demonstrated that the integration of CS and DL can potentially lead to a promising future direction for MRI reconstruction.

\subsection{Advancements and impact}

DL-based MRI reconstruction methods have demonstrated remarkable capabilities in accelerating MRI acquisition without compromising image quality. Techniques such as U-nets, GANs, transformers, and diffusion models have shown superior performance in reconstructing high-quality images from undersampled \textit{k}-space data. These methods leverage neural networks to learn complex mappings from undersampled to fully sampled data, enabling faster and more accurate image reconstructions compared to traditional CS methods.

Although DL is distinct from CS, certain DL approaches integrate CS-inspired principles. For instance, unrolled networks convert iterative optimization steps in traditional CS algorithms into trainable neural network layers, pairing the interpretability of CS with the efficiency of DL. Data consistency layers further elevate reconstruction fidelity by enforcing adherence to the acquired \textit{k}-space data. These hybrid methods underscore the potential of merging CS principles with DL frameworks, yielding enhanced robustness and generalization across varied clinical scenarios.

\subsection{The Role of Multi-Coil Data in MRI Reconstruction}

Multi-coil data play a pivotal role in modern MRI reconstruction, capitalizing on spatial sensitivity profiles from multiple coils to enhance reconstruction accuracy and expedite acquisitions. Most studies examined in this review evaluated their methods on both single-coil and multi-coil datasets. For example, resources such as fastMRI provide raw \textit{k}-space data for multi-coil acquisitions while offering single-coil reconstructions for more streamlined evaluations. This dual-validation approach bolsters robustness across acquisition settings and allows direct comparisons with established methods.

Despite widespread adoption, challenges remain in fully leveraging multi-coil data for realistic demonstrations. A major hurdle is the significant computational demand associated with estimating coil sensitivity maps--often obtained through the ESPIRiT algorithm--which can hinder real-time clinical applications. Future research should focus on integrating sensitivity estimation within reconstruction workflows, especially for DL-based methods. Additionally, the limited availability of comprehensive multi-coil datasets that encompass a variety of anatomical regions and pathologies poses a barrier to further advancements in this area.

\subsection{Limitations and challenges}

Integrating DL with CS in MRI has shown significant promise in enhancing image quality and reducing acquisition times. This innovative approach holds great potential for revolutionizing clinical practice. However, several challenges must be addressed to fully realize this potential.

A primary limitation lies in the high computational load that characterizes many DL models. Given their reliance on GPU-accelerated hardware and substantial memory resources, these models may be inaccessible in cost-constrained settings. Developing more efficient algorithms that preserve high performance while reducing computational demands represents a vital step toward broader clinical adoption.

Data availability and quality also pose critical concerns. Large, high-quality datasets are essential for training robust DL models, yet many institutions cannot access extensive databases. Although resources like the fastMRI dataset have been invaluable, they primarily offer multi-coil data that can be converted into single-coil reconstructions. Moreover, the shortage of raw 3D and 4D \textit{k}-space data with pathologies constrains model generalization to different clinical contexts.

Another key challenge is achieving model generalizability. DL models trained on specific datasets may struggle to perform well on data from different institutions or with varying imaging protocols. This lack of robustness is problematic, especially when dealing with diverse patient anatomies and pathological conditions. Ensuring that models are adaptable and perform consistently across various scenarios remains a complex task. Overfitting to specific datasets and the difficulty in generalizing to new data are ongoing issues that need to be addressed.

Artifacts prevalent in DL-generated MRI data include synthetic structures or ``hallucinations,'' which can arise at high acceleration factors, as well as blurring or smoothing artifacts that obscure critical details. Incoherent aliasing may also emerge from suboptimal under-sampling or model limitations, yielding misleading features that compromise diagnostic integrity. Such artifacts hinder clinical efficacy by masking true pathologies or introducing fictitious anatomical features.

Addressing these challenges requires a multidisciplinary approach, involving collaboration between researchers, clinicians, and regulatory agencies. By focusing on model generalizability, workflow integration, computational efficiency, and regulatory compliance, the deployment of DL-based MRI reconstruction models in clinical practice can be effectively realized, leading to improved patient outcomes and enhanced diagnostic capabilities.

\subsection{Future directions}

The objective of DL-based MRI reconstruction approaches is to decrease imaging time, thereby enhancing throughput and minimizing the chances of patient movement. Apart from this, other methods, such as super-resolution and image synthesis, can also be adopted to decrease imaging time. These approaches strive to enhance the resolution of images obtained from lower spatial resolution~\cite{chang_high-resolution_2024,xie_synthesizing_2022} and produce MRI images of high spatial resolution and SNR from lower $ B_0 $~\cite{eidex_high-resolution_2023}. These techniques can markedly improve MRI imaging with permanent magnets having low $ B_0 $, ultimately increasing the image quality. Portable MRI scanners widely use permanent magnets, which substantially reduce maintenance costs and make them more suitable for low-income and middle-income countries. Although these techniques can reduce the need for stronger gradient magnets and minimize patients’ nerve stimulation, they were not included in this systematic review since they did not employ compressed sensing algorithms to train their models.

The coil sensitivity map shown in Figure~\ref{fig:pi_coil_senmap}b is essential to consider the spatially varying sensitivity of the receiver coils. A precise sensitivity map is crucial for generating consistent and accurate MRI images and quantitative MRI maps across different hospitals. Most of the reviewed studies use the ESPIRiT algorithm~\cite{uecker_espiriteigenvalue_2014} provided by the Berkeley Advanced Reconstruction Toolbox (\url{https://mrirecon.github.io/bart/}) to pre-calculate the sensitivity map. However, this algorithm involves significant computation that may hinder its application in MRI-guided surgery and treatment, where rapid image reconstruction is required. We anticipate that DL-based MRI reconstruction techniques, which can simultaneously predict coil sensitivity maps and coil images~\cite{feng_imjense_2023,sun_joint_2023,wang_faithful_2024, https://doi.org/10.1002/mrm.30121}, will be explored further in the future, particularly for MRI-guided treatment methods such as MRI-guided adaptive radiation therapy. Furthermore, while most MRI reconstruction methods require prior knowledge of the sampling pattern, recent advancements have developed techniques to predict or optimize sampling patterns using DL~\cite{aggarwal_j-modl_2020,chung_learning_based_2019,seo_simultaneously_2022} may further expedite and improve reconstruction efficiency.

\paragraph{Datasets}
While public datasets like fastMRI have significantly advanced MRI reconstruction research, they primarily offer multi-coil data with options for derived single-coil reconstructions. Future expansions of these datasets to include broader anatomical coverage, varied acquisition protocols, and a wider range of pathologies would markedly amplify both realism and applicability. Emphasizing the release of raw multi-coil \textit{k}-space data is particularly important for enabling more robust evaluations and ensuring clinical relevance. By encompassing greater diversity, future datasets will help address the pressing challenges of model generalization and artifact management.

Future works also involve designing DL-based reconstruction methods that specifically target artifact suppression and detection. Hybrid networks that fuse DL with traditional CS techniques and data-consistency layers could offer crucial improvements, mitigating the synthetic artifacts often introduced by purely data-driven approaches. Self-supervised learning offers another pathway to fortify model robustness, allowing adaptation across heterogeneous scanner types and patient populations with minimal labeled data. FL frameworks can further facilitate model training on distributed datasets, potentially alleviating biases linked to limited institutional data. Introducing dedicated artifact detection and correction modules may also refine image quality, enhancing diagnostic precision prior to clinical assessment.

The implementation of strict clinical outcome metrics is vital for future research. Relying too heavily on metrics like SSIM and PSNR can be misleading if they fail to capture important clinically relevant features. Standardizing evaluation measures that are clinically meaningful will help ensure that improvements in reconstruction fidelity lead to real benefits in patient care and diagnosis.

\paragraph{Vision-Large Models}
In a recent systematic review by Hu, Mingzhe, \textit{et al.}, the use of large language models (LLMs) in medical imaging was explored in detail~\cite{hu_advancing_2024}. Although a separate study indicated that a vision-and-language model (VLM) demonstrated benefits in computed tomography reconstruction~\cite{xing2024deep}, our comprehensive review has identified that the utilization of vision and language model (VLMs) in training DL-based MRI reconstruction is currently quite limited. We propose that VLMs could have a significant impact on DL-based MRI reconstruction. By leveraging insights from extensive textual data, LLMs can provide context-aware constraints to enhance data consistency and regularization, and tailor reconstruction models to specific MRI sequences or anatomical regions. Additionally, their ability to generalize knowledge across different imaging domains and patient populations offers potential solutions for challenges related to out-of-distribution or out-of-region data. Therefore, incorporating LLMs into DL-based reconstruction frameworks could accelerate model development, enhance clinical interpretability, and ultimately lead to more robust and efficient MRI reconstructions.

\section{Concluding remarks}

In conclusion, DL-based MRI reconstruction approach has significantly advanced MRI reconstruction by reducing acquisition times and improving image quality. Addressing the challenges of computational demands and data variability is essential for broader clinical adoption. Future research should explore FL and the VLMs robustness and clinical applicability.

\section*{Acknowledgments}
This research is supported in part by the National Institutes of Health under Award Numbers R56EB033332, R01EB032680, and R01CA272991.

\section*{Conflicts of interest}
There are no conflicts of interest declared by the authors.


\end{document}